\documentclass{article}

\usepackage{arxiv}

\usepackage[utf8]{inputenc} 
\usepackage[T1]{fontenc}    
\usepackage{hyperref}       
\usepackage{url}            
\usepackage{booktabs}       
\usepackage{amsfonts}       
\usepackage{nicefrac}       
\usepackage{microtype}      
\usepackage{lipsum}
\usepackage{graphicx}
\graphicspath{ {./images/} }

\usepackage{natbib}

\usepackage{amsmath}
\usepackage{multirow}

\usepackage{todonotes}
\usepackage{xcolor}
\RequirePackage{tikz}
\usetikzlibrary{arrows,arrows.meta, shapes, calc, positioning, patterns, patterns.meta}
\usepackage{minted}

\definecolor{rosewater}{HTML}{DC8A78}
\definecolor{flamingo}{HTML}{DD7878}
\definecolor{pink}{HTML}{EA76CB}
\definecolor{mauve}{HTML}{8839EF}
\definecolor{red}{HTML}{D20F39}
\definecolor{maroon}{HTML}{E64553}
\definecolor{peach}{HTML}{FE640B}
\definecolor{yellow}{HTML}{DF8E1D}
\definecolor{green}{HTML}{40A02B}
\definecolor{teal}{HTML}{179299}
\definecolor{sky}{HTML}{04A5E5}
\definecolor{sapphire}{HTML}{209FB5}
\definecolor{blue}{HTML}{1E66F5}
\definecolor{lavender}{HTML}{7287FD}
\definecolor{text}{HTML}{4C4F69}
\definecolor{subtext1}{HTML}{5C5F77}
\definecolor{subtext0}{HTML}{6C6F85}
\definecolor{overlay2}{HTML}{7C7F93}
\definecolor{overlay1}{HTML}{8C8FA1}
\definecolor{overlay0}{HTML}{9CA0B0}
\definecolor{surface2}{HTML}{ACB0BE}
\definecolor{surface1}{HTML}{BCC0CC}
\definecolor{surface0}{HTML}{CCD0DA}
\definecolor{base}{HTML}{EFF1F5}
\definecolor{mantle}{HTML}{E6E9EF}
\definecolor{crust}{HTML}{DCE0E8}

\title{Optimizing Fairness in Production Planning: A Human-Centric Approach to Machine and Workforce Allocation}

\author{
 Alexander Nasuta \\
  WZL-IQS\\
  RWTH Aachen University\\
  \\
  \texttt{alexander.nasuta@wzl-iqs.rwth-aachen.de} \\
   \And
  Alessandro Cisi \\
  Centro Ricerche Fiat\\ \\ \\ \\
  \And
Sylwia Olbrych \\
  WZL-IQS\\
  RWTH Aachen University\\ \\ \\
    \And
Gustavo Vieira \\
  MORE\\
  Laboratório Colaborativo Montanhas de Investigação\\ \\ \\
    \And
Rui Fernandes \\
   MORE\\
  Laboratório Colaborativo Montanhas de Investigação\\ \\ \\
    \And
Lucas Paletta \\
  Joanneum Research Forschungsgesellschaft mbH\\ \\ \\ \\
    \And
Marlene Mayr \\
 BOC Products \& Services AG\\ \\ \\ \\
    \And
Rishyank Chevuri \\
  Jotne Connect AS\\ \\ \\ \\
    \And
Robert Woitsch \\
  BOC Products \& Services AG\\ \\ \\ \\
    \And
Hans Aoyang Zhou \\
  WZL-IQS\\
  RWTH Aachen University\\ \\ \\
    \And
Anas Abdelrazeq \\
  WZL-IQS\\
  RWTH Aachen University\\ \\ \\
    \And
Robert H. Schmitt \\
  WZL-IQS\\
  RWTH Aachen University\\ \\ \\
}

\begin{document}
\maketitle
\begin{abstract}
This work presents a two-layer, human-centric production planning framework designed to optimize both operational efficiency and workforce fairness in industrial manufacturing.
The first layer formulates the Order–Line allocation as a Constraint Programming (CP) problem, generating high-utilization production schedules that respect machine capacities, processing times, and due dates. 
The second layer models Worker–Line allocation as a Markov Decision Process (MDP), integrating human factors such as worker preference, experience, resilience, and medical constraints into the assignment process.
Three solution strategies greedy allocation, MCTS, and RL are implemented and compared across multiple evaluation scenarios.  
The proposed system is validated through 16 test sessions with domain experts from the automotive industry, combining quantitative key performance indicators (KPIs) with expert ratings. 
Results indicate that the CP-based scheduling approach produces compact, feasible production plans with low tardiness, while the MDP-based worker allocation significantly improves fairness and preference alignment compared to baseline approaches. 
Domain experts rated both the Order–Line and Worker–Line components as effective and highlighted opportunities to further refine the objective function to penalize excessive earliness and improve continuity in worker assignments.  
Overall, the findings demonstrate that combining CP with learning-based decision-making provides a robust approach for human-centric production planning. 
The approach enables simultaneous optimization of throughput and workforce well-being, offering a practical foundation for fair and efficient manufacturing scheduling in industrial settings.
\end{abstract}


\section{Introduction}
\label{sec:intro}

Production planning in industrial environments is a multifaceted optimization problem involving the allocation of tasks to machines and the scheduling of personnel. 
Historically, such processes have prioritized productivity metrics such as throughput, machine utilization, and on-time delivery \citep{Jaehn2019}. 
However, there is a growing emphasis on human-centric planning approaches that account for worker well-being, preferences, and fairness \citep{katiraee2021considerationReview}. 
This shift reflects the increasing recognition that sustainable production depends not only on operational efficiency but also on equitable workload distribution, which in turn contributes to lower employee turnover and improved long-term workforce stability. 
Our use case is situated in the automotive sector, where various components, referred to as geometries, such as car doors or hoods are manufactured on specialized machines, also referred to as line in the use case. 
These tasks vary significantly in their physical demands, with some geometries requiring substantial manual effort. This variability has led to fairness concerns, particularly when certain workers are disproportionately assigned to more physically demanding tasks. 
Moreover, machine operation requires specific training, and workers differ in their resilience, experience, and task preferences. 
These human factors must therefore be integrated into the production planning process. 
The problem consists of two interdependent sub-problems: assigning geometries to machines over time (i.e., production scheduling), and allocating workers to machines while considering their constraints and preferences. 
The allocation of products to machines in the use case corresponds to the flexible job shop scheduling problem (FJSP), a well-known NP-hard combinatorial optimization problem that generalizes the classical job shop problem by allowing operations to be processed on multiple alternative machines (cf. \cite{aymenDiss}). 
Introducing the additional constraints for worker allocation based on personal attributes significantly increases the problem complexity. 
Due to the intractability of exact methods for large-scale instances, heuristic and metaheuristic approaches are commonly employed to approximate high-quality solutions. 
In recent years, learning-based methods such as Reinforcement Learning (RL) and neural Monte Carlo Tree Search (neural MCTS) have emerged as promising tools for complex scheduling tasks \citep{marcoDiss}. 
These approaches learn adaptive scheduling strategies through interaction with simulated environments, thereby coping with the combinatorial complexity and multi-objective trade-offs inherent in human-centric production planning. 

To support decision-makers in addressing these challenges, we propose a service-based system that allows users to select from multiple algorithmic strategies and to customize cost functions through parameterization of preference, experience, resilience. 
This flexibility enables the exploration of alternative scheduling scenarios and facilitates fairer planning outcomes. 
The system thus empowers production line managers to explicitly account for human factors in their decision-making processes.

The remainder of this article is structured as follows: 
Section~\ref{sec:related-work} provides a brief review of related work. 
Section~\ref{sec:Methodology} presents the methodology and describes how human factors were integrated into our production planning use case. 
Evaluation results based on expert feedback are presented and discussed in Section~\ref{sec:results_and_discussion}, followed by conclusions in Section~\ref{sec:conclusion}.

\section{Related Work}
\label{sec:related-work}

Production planning and workforce scheduling in manufacturing environments have increasingly incorporated human factors to address challenges arising from workforce diversity, physical demands, and evolving technological contexts. 
Recent studies highlight the need for integrating worker-specific factors, preferences, and well-being into scheduling models to enhance both productivity and fairness. 
Additionally, advanced algorithmic approaches, including RL and MCTS methods, have been investigated for their potential to solve complex allocation problems in dynamic production settings. 
The following paragraphs review briefly relevant literature on human factors in manufacturing, their incorporation into production scheduling, and emerging solution techniques.

The demographic shift towards an aging workforce poses significant challenges to manufacturing systems, necessitating adaptations that consider physical, cognitive, and ergonomic needs. \cite{alves2024sociodemographic} systematically review this issue within the framework of Industry 4.0 and the nascent Industry 5.0 paradigm, advocating for human-centric production designs that sustain worker health and system resilience. 
Their analysis categorizes research into ageing, technology, human factors, and ergonomics, and emphasizes the limited real-world adoption of such integrative approaches. 
Complementing this perspective, \cite{finco2020considering} develop a job rotation scheduling model that incorporates age-related physical capacity and experience differences, demonstrating that human-centric constraints can improve workload distribution without expensive infrastructure changes. 
Empirical surveys by \cite{jeon2016preferred} support the benefits of job rotation for worker satisfaction and musculoskeletal disorder prevention, though effectiveness varies by age group. 
Collectively, these studies underscore the importance of ergonomic and well-being considerations in sustaining productive and inclusive manufacturing workforces.
\cite{katiraee2021considerationReview} conducted a systematic literature review focusing on how differences among workers, such as skills, age, gender, and anthropometric characteristics, affect production system performance and should be considered in system modelling and design. 
Their work identifies that, while skill differences are frequently addressed, other human factors like age and physical capabilities remain underexplored. 
They advocate for more individualized modelling approaches, use of real-world ergonomic and performance data, and adaptive workstation design to improve health, job satisfaction, and productivity in heterogeneous workforces.
Wearable-based human sensors have become increasingly relevant in production environments \citep{rauch2020anthropocentric}, as they enable continuous monitoring of physiological and behavioral states to assess workload, stress, and fatigue. 
Systematic reviews highlight their potential to improve occupational safety, productivity, and well-being by providing objective, real-time insights into human factors \citep{paletta2021towards, lu2022outlook}. 
Broad syntheses further emphasize that such technologies contribute to adaptive production systems by integrating human-centered data into decision-making processes \citep{paletta2023digital, paletta2024resilience}.

Recent literature reveals a growing emphasis on embedding human-centered principles within production scheduling algorithms. 
\cite{burgert2024workforce} conduct a comprehensive review of scheduling models, evaluating their alignment with human-centered work design factors such as preferences, qualifications, and ergonomics. 
While skill matching and task variety are sometimes considered, dimensions like autonomy and social interaction are often overlooked. 
To address this, the authors propose a conceptual model that balances task allocation fairness with worker preferences, advocating for data-driven quantification of subjective well-being metrics. 
\cite{katiraee2021consideration} present a bi-objective line balancing and worker assignment model that incorporates workers’ self-assessed physical effort through a novel Worker Task Categorization Matrix. 
Their approach balances cycle time efficiency with ergonomic constraints, highlighting risks associated with assigning physically unsuitable tasks. 
In a healthcare context, \cite{gerlach2024exploring} analyze nurse perceptions of AI-based scheduling systems, finding that fairness, transparency, and human oversight are critical to acceptance. 
These works collectively indicate that effective production scheduling must integrate multifaceted human factors to support equitable and sustainable workforce management.

To address the complexity of production scheduling and workforce allocation, various computational methods have been explored. 
\cite{kemmerling2024beyond} review the application of neural MCTS beyond its original use in board games, highlighting its adaptation to deterministic single-agent environments with large action spaces, including scheduling problems.
Their study identifies customization challenges and emphasizes the need for modular frameworks and domain-specific tuning to optimize search performance. 
\cite{waubert2022reliability} survey the use of RL for production scheduling, focusing on the critical aspect of reliability in industrial applications. 
They note a disparity between advanced DRL techniques and practical requirements, proposing standardized evaluation criteria to improve robustness and applicability. 
Both reviews highlight the potential of AI-based methods to enhance scheduling efficiency and adaptability, provided that human factors and real-world constraints are sufficiently incorporated. 
Traditional approaches, including CP and heuristics, remain relevant for their interpretability and ease of integration with domain knowledge.

\section{Methodology}
\label{sec:Methodology}

This section outlines the modeling of the use case, the selected solution approaches, the system architecture, and the evaluation setup. 
The overall objective is to develop a human‐centric production planning system capable of optimizing both fairness and operational efficiency in the allocation of machines and workers. 
The problem is formalized in two sequential layers: (1) allocation of product orders to machines over time, and (2) allocation of workers to machines considering individual attributes, constraints, and preferences.

This chapter first elaborates on the rationale behind adopting a two-layered approach to the allocation problem. 
The subsequent sections present the core concepts underlying each layer, followed by their concrete realization in the given use case. 
The first layer is based on CP, which is introduced as the fundamental modeling paradigm for allocating product orders to machines, and the problem is formally defined as a CP program. 
The second layer builds on a Markov Decision Process (MDP), which provides the framework for allocating workers to machines while incorporating human-centric factors such as preferences, resilience, experience, and medical constraints. 
The integration of both approaches into a unified planning system is then detailed in the system architecture section. 
Finally, the chapter outlines the evaluation strategy used to assess the effectiveness of the proposed two-layer solution with respect to both fairness and operational efficiency.


\subsection{Problem Definition}

The use case involves assigning workers to metal‐processing machines (referred to as lines) to produce batches of products (referred to as geometries). 
Production on each line is physically demanding for the personnel operating the machine, requires machine-specific training, and must comply with medical suitability constraints.
Furthermore, workers differ in their experience levels, physical resilience, and expressed preferences for specific geometry–line combinations. The planning problem involves two mains aspects:

\begin{enumerate}
    \item \textbf{Order–Line Allocation}: Assigning geometries to machines over a given time horizon, subject to setup and processing times, machine availability, and order deadlines. 
    In this use case, the scheduling task corresponds to a variant of the Flexible Job Shop Scheduling Problem (FJSP), where each operation must be executed on one of several alternative machines.
    \item \textbf{Worker–Line Allocation}:
    Assigning available workers to the scheduled machine operations, subject to medical restrictions, experience levels, resilience scores, and preference values. 
    The objective is to maximize average worker preference, resilience, and experience and to allocate the workers such that the distribution of preferred and less preferred tasks among workers is as equal as possible.
    An even distribution of preferred and unpreferred tasks is considered fair in this use case.
\end{enumerate}

The initial prototype formulated the worker allocation as a single-shift linear sum assignment problem and solved it using the Hungarian algorithm, utilizing Google OR-Tools \citep{ortools} implementation of the algorithm.
This was computationally efficient for small, single‐shift cases but became intractable when integrated directly into the Order–Line CP model for multi‐day horizons. 
Early experiments showed that joint solving frequently failed to return feasible solutions withing practical time horizons.
Consequently, the two main aspects are solved individually.
This separation reduced computational complexity and allowed each layer to be optimized with methods suited to its characteristics.

A high Level concept of the approach is visualized in Figure 1.
First the production schedule is genrated based on the order and  spec sheets of the available machines.
Afterwards the resulting production schedule is taken alongside shift plan for personal and preference, resillience, experiance, and medical condition (ability to work on the machine) data to find a fair allocation of the workforce to the machines.
These values are referred to as Human Factors.
For each geometry contained in an order, a dedicated lane is shown on the y-axis of the Gantt charts. 
Within each lane, a rectangle represents the planned production interval, with its start and end points corresponding to the scheduled times. 
The rectangle color encodes the line on which the geometry is processed. The x-axis denotes time, structured into three daily shifts (06:00–14:00, 14:00–22:00, and 22:00–06:00). 
No production is scheduled between Saturday 06:00 and Monday 06:00, reflecting weekend downtime. The optimization algorithms operate in a minute-based time domain starting from a given reference point, while weekends are excluded from the computational horizon. 
For example, if the reference time is Saturday 05:59 (defined as 0 minutes in solver time), the following Monday 06:01 corresponds to 2 minutes in solver time.
Even though all optimization is performed in this solver-specific minute-based domain, actual clock times are added in all figures to support readability.
The resulting schedule is ultimately transformed back into calendar time before being passed to the user.

\definecolor{chart-red}{RGB}{230, 25, 75}       
\definecolor{chart-green}{RGB}{60, 180, 75}     
\definecolor{chart-blue}{RGB}{0, 130, 200}      
\definecolor{chart-orange}{RGB}{245, 130, 48}   
\definecolor{chart-purple}{RGB}{145, 30, 180}   
\definecolor{chart-cyan}{RGB}{70, 240, 240}     
\definecolor{chart-magenta}{RGB}{240, 50, 230}  
\definecolor{chart-lime}{RGB}{210, 245, 60}     
\definecolor{chart-pink}{RGB}{250, 190, 190}    
\definecolor{chart-teal}{RGB}{0, 128, 128}      
\definecolor{chart-lavender}{RGB}{230, 190, 255} 
\definecolor{chart-brown}{RGB}{170, 110, 40}    
\definecolor{chart-beige}{RGB}{255, 250, 200}   
\definecolor{chart-maroon}{RGB}{128, 0, 0}      
\definecolor{chart-mint}{RGB}{170, 255, 195}    

\definecolor{chart-olive}{RGB}{128, 128, 0}     
\definecolor{chart-apricot}{RGB}{255, 215, 180} 
\definecolor{chart-navy}{RGB}{0, 0, 128}        
\definecolor{chart-gray}{RGB}{128, 128, 128}    

\definecolor{css-AliceBlue}{RGB}{240,248,255}
\definecolor{css-AntiqueWhite}{RGB}{250,235,215}
\definecolor{css-Aqua}{RGB}{0,255,255}
\definecolor{css-Aquamarine}{RGB}{127,255,212}
\definecolor{css-Azure}{RGB}{240,255,255}
\definecolor{css-Beige}{RGB}{245,245,220}
\definecolor{css-Bisque}{RGB}{255,228,196}
\definecolor{css-Black}{RGB}{0,0,0}
\definecolor{css-BlanchedAlmond}{RGB}{255,235,205}
\definecolor{css-Blue}{RGB}{0,0,255}
\definecolor{css-BlueViolet}{RGB}{138,43,226}
\definecolor{css-Brown}{RGB}{165,42,42}
\definecolor{css-BurlyWood}{RGB}{222,184,135}
\definecolor{css-CadetBlue}{RGB}{95,158,160}
\definecolor{css-Chartreuse}{RGB}{127,255,0}
\definecolor{css-Chocolate}{RGB}{210,105,30}
\definecolor{css-Coral}{RGB}{255,127,80}
\definecolor{css-CornflowerBlue}{RGB}{100,149,237}
\definecolor{css-Cornsilk}{RGB}{255,248,220}
\definecolor{css-Crimson}{RGB}{220,20,60}
\definecolor{css-Cyan}{RGB}{0,255,255}
\definecolor{css-DarkBlue}{RGB}{0,0,139}
\definecolor{css-DarkCyan}{RGB}{0,139,139}
\definecolor{css-DarkGoldenRod}{RGB}{184,134,11}
\definecolor{css-DarkGray}{RGB}{169,169,169}
\definecolor{css-DarkGrey}{RGB}{169,169,169}
\definecolor{css-DarkGreen}{RGB}{0,100,0}
\definecolor{css-DarkKhaki}{RGB}{189,183,107}
\definecolor{css-DarkMagenta}{RGB}{139,0,139}
\definecolor{css-DarkOliveGreen}{RGB}{85,107,47}
\definecolor{css-DarkOrange}{RGB}{255,140,0}

\definecolor{css-DarkOrchid}{RGB}{153,50,204}
\definecolor{css-DarkRed}{RGB}{139,0,0}
\definecolor{css-DarkSalmon}{RGB}{233,150,122}
\definecolor{css-DarkSeaGreen}{RGB}{143,188,143}
\definecolor{css-DarkSlateBlue}{RGB}{72,61,139}
\definecolor{css-DarkSlateGray}{RGB}{47,79,79}
\definecolor{css-DarkSlateGrey}{RGB}{47,79,79}
\definecolor{css-DarkTurquoise}{RGB}{0,206,209}
\definecolor{css-DarkViolet}{RGB}{148,0,211}
\definecolor{css-DeepPink}{RGB}{255,20,147}
\definecolor{css-DeepSkyBlue}{RGB}{0,191,255}
\definecolor{css-DimGray}{RGB}{105,105,105}
\definecolor{css-DimGrey}{RGB}{105,105,105}
\definecolor{css-DodgerBlue}{RGB}{30,144,255}
\definecolor{css-FireBrick}{RGB}{178,34,34}
\definecolor{css-FloralWhite}{RGB}{255,250,240}
\definecolor{css-ForestGreen}{RGB}{34,139,34}
\definecolor{css-Fuchsia}{RGB}{255,0,255}
\definecolor{css-Gainsboro}{RGB}{220,220,220}
\definecolor{css-GhostWhite}{RGB}{248,248,255}
\definecolor{css-Gold}{RGB}{255,215,0}
\definecolor{css-GoldenRod}{RGB}{218,165,32}
\definecolor{css-Gray}{RGB}{128,128,128}
\definecolor{css-Grey}{RGB}{128,128,128}
\definecolor{css-Green}{RGB}{0,128,0}
\definecolor{css-GreenYellow}{RGB}{173,255,47}
\definecolor{css-HoneyDew}{RGB}{240,255,240}
\definecolor{css-HotPink}{RGB}{255,105,180}
\definecolor{css-IndianRed}{RGB}{205,92,92}
\definecolor{css-Indigo}{RGB}{75,0,130}
\definecolor{css-Ivory}{RGB}{255,255,240}
\definecolor{css-Khaki}{RGB}{240,230,140}
\definecolor{css-Lavender}{RGB}{230,230,250}
\definecolor{css-LavenderBlush}{RGB}{255,240,245}
\definecolor{css-LawnGreen}{RGB}{124,252,0}
\definecolor{css-LemonChiffon}{RGB}{255,250,205}
\definecolor{css-LightBlue}{RGB}{173,216,230}
\definecolor{css-LightCoral}{RGB}{240,128,128}
\definecolor{css-LightCyan}{RGB}{224,255,255}
\definecolor{css-LightGoldenRodYellow}{RGB}{250,250,210}
\definecolor{css-LightGray}{RGB}{211,211,211}
\definecolor{css-LightGrey}{RGB}{211,211,211}
\definecolor{css-LightGreen}{RGB}{144,238,144}
\definecolor{css-LightPink}{RGB}{255,182,193}
\definecolor{css-LightSalmon}{RGB}{255,160,122}
\definecolor{css-LightSeaGreen}{RGB}{32,178,170}
\definecolor{css-LightSkyBlue}{RGB}{135,206,250}
\definecolor{css-LightSlateGray}{RGB}{119,136,153}
\definecolor{css-LightSlateGrey}{RGB}{119,136,153}
\definecolor{css-LightSteelBlue}{RGB}{176,196,222}
\definecolor{css-LightYellow}{RGB}{255,255,224}
\definecolor{css-Lime}{RGB}{0,255,0}
\definecolor{css-LimeGreen}{RGB}{50,205,50}
\definecolor{css-Linen}{RGB}{250,240,230}
\definecolor{css-Magenta}{RGB}{255,0,255}
\definecolor{css-Maroon}{RGB}{128,0,0}
\definecolor{css-MediumAquaMarine}{RGB}{102,205,170}
\definecolor{css-MediumBlue}{RGB}{0,0,205}
\definecolor{css-MediumOrchid}{RGB}{186,85,211}
\definecolor{css-MediumPurple}{RGB}{147,112,219}
\definecolor{css-MediumSeaGreen}{RGB}{60,179,113}
\definecolor{css-MediumSlateBlue}{RGB}{123,104,238}
\definecolor{css-MediumSpringGreen}{RGB}{0,250,154}
\definecolor{css-MediumTurquoise}{RGB}{72,209,204}
\definecolor{css-MediumVioletRed}{RGB}{199,21,133}
\definecolor{css-MidnightBlue}{RGB}{25,25,112}
\definecolor{css-MintCream}{RGB}{245,255,250}
\definecolor{css-MistyRose}{RGB}{255,228,225}
\definecolor{css-Moccasin}{RGB}{255,228,181}
\definecolor{css-NavajoWhite}{RGB}{255,222,173}
\definecolor{css-Navy}{RGB}{0,0,128}
\definecolor{css-OldLace}{RGB}{253,245,230}
\definecolor{css-Olive}{RGB}{128,128,0}
\definecolor{css-OliveDrab}{RGB}{107,142,35}
\definecolor{css-Orange}{RGB}{255,165,0}
\definecolor{css-OrangeRed}{RGB}{255,69,0}
\definecolor{css-Orchid}{RGB}{218,112,214}
\definecolor{css-PaleGoldenRod}{RGB}{238,232,170}
\definecolor{css-PaleGreen}{RGB}{152,251,152}
\definecolor{css-PaleTurquoise}{RGB}{175,238,238}
\definecolor{css-PaleVioletRed}{RGB}{219,112,147}
\definecolor{css-PapayaWhip}{RGB}{255,239,213}
\definecolor{css-PeachPuff}{RGB}{255,218,185}
\definecolor{css-Peru}{RGB}{205,133,63}
\definecolor{css-Pink}{RGB}{255,192,203}
\definecolor{css-Plum}{RGB}{221,160,221}
\definecolor{css-PowderBlue}{RGB}{176,224,230}
\definecolor{css-Purple}{RGB}{128,0,128}
\definecolor{css-RebeccaPurple}{RGB}{102,51,153}
\definecolor{css-Red}{RGB}{255,0,0}
\definecolor{css-RosyBrown}{RGB}{188,143,143}
\definecolor{css-RoyalBlue}{RGB}{65,105,225}
\definecolor{css-SaddleBrown}{RGB}{139,69,19}
\definecolor{css-Salmon}{RGB}{250,128,114}
\definecolor{css-SandyBrown}{RGB}{244,164,96}
\definecolor{css-SeaGreen}{RGB}{46,139,87}
\definecolor{css-SeaShell}{RGB}{255,245,238}
\definecolor{css-Sienna}{RGB}{160,82,45}
\definecolor{css-Silver}{RGB}{192,192,192}
\definecolor{css-SkyBlue}{RGB}{135,206,235}
\definecolor{css-SlateBlue}{RGB}{106,90,205}
\definecolor{css-SlateGray}{RGB}{112,128,144}
\definecolor{css-SlateGrey}{RGB}{112,128,144}
\definecolor{css-Snow}{RGB}{255,250,250}
\definecolor{css-SpringGreen}{RGB}{0,255,127}
\definecolor{css-SteelBlue}{RGB}{70,130,180}
\definecolor{css-Tan}{RGB}{210,180,140}
\definecolor{css-Teal}{RGB}{0,128,128}
\definecolor{css-Thistle}{RGB}{216,191,216}
\definecolor{css-Tomato}{RGB}{255,99,71}
\definecolor{css-Turquoise}{RGB}{64,224,208}
\definecolor{css-Violet}{RGB}{238,130,238}
\definecolor{css-Wheat}{RGB}{245,222,179}
\definecolor{css-White}{RGB}{255,255,255}
\definecolor{css-WhiteSmoke}{RGB}{245,245,245}
\definecolor{css-Yellow}{RGB}{255,255,0}
\definecolor{css-YellowGreen}{RGB}{154,205,50}

\begin{figure}[htb]
\centerline{
\begin{tikzpicture}

\draw[fill=base] (0,2) rectangle ++(2.25,1) node[midway, align=center] {Static Data};
\draw[fill=base] (2.75,2) rectangle ++(2.25,1) node[midway, align=center] {Dynamic Data};

\draw[-latex] (1.125, 2) -- ++(0, -0.75);
\draw[-latex] (3.875, 2) -- ++(0, -0.75);

\draw[fill=base] (0,-0.25) rectangle ++(5,1.5) node[midway, align=center] {Order–Line Allocation \\ (Constraint Programming)};

\draw[-latex] (2.5, -0.25) -- ++(0, -0.75);

\begin{scope}[shift={(0,-3.5)}, scale=0.125]
    \draw[-latex] (0,19) node[above] {} -- (0,0) --  (44,0);

    \foreach \x/\xtext in {4, 12,...,40}
    \draw[shift={(\x,0)}] (0,0) -- (0,-0.75);
    
    \foreach \x/\xtext in {0, 8,...,40}
    \draw[shift={(\x,0)}] (0,0) -- (0,-1);

    \node[align=left, anchor=west] at (-10, 16) {\tiny Geometry 8};
    \node[align=left, anchor=west] at (-10, 14) {\tiny Geometry 7};
    \node[align=left, anchor=west] at (-10, 12) {\tiny Geometry 6};
    \node[align=left, anchor=west] at (-10, 10) {\tiny Geometry 5};
    \node[align=left, anchor=west] at (-10, 8) {\tiny Geometry 4};
    \node[align=left, anchor=west] at (-10, 6) {\tiny Geometry 3};
    \node[align=left, anchor=west] at (-10, 4) {\tiny Geometry 2};
    \node[align=left, anchor=west] at (-10, 2) {\tiny Geometry 1};
    
    \node[align=left, anchor=west] at (45,0) {\scriptsize time};
    \node[align=center, anchor=north, yshift=0.1mm] at (48.5,-1) {\tiny [min]};
    \node[align=center, anchor=north, yshift=0.1mm] at (48.5,-3) {\tiny [Day hh:mm]};
    
    \node[align=center, anchor=north] at (0,-1) {\tiny 0};
    \node[align=center, anchor=north] at (8,-1) {\tiny 480};
    \node[align=center, anchor=north] at (16,-1) {\tiny 960};
    \node[align=center, anchor=north] at (24,-1) {\tiny 1440};
    \node[align=center, anchor=north] at (32,-1) {\tiny 1920};
    \node[align=center, anchor=north] at (40,-1) {\tiny 2400};

    \node[align=center, anchor=north] at (0,-3) {\tiny Mon 6:00};
    \node[align=center, anchor=north] at (8,-3) {\tiny Mon 14:00};
    \node[align=center, anchor=north] at (16,-3) {\tiny Mon 22:00};
    \node[align=center, anchor=north] at (24,-3) {\tiny Tue 6:00};
    \node[align=center, anchor=north] at (32,-3) {\tiny Tue 14:00};
    \node[align=center, anchor=north] at (40,-3) {\tiny Tue 22:00};

    \draw[draw=text, fill=sapphire] (0, 3) rectangle (8, 5);

    \draw[draw=text, fill=sapphire] (8, 5) rectangle (20, 7);

   
    \draw[draw=text, fill=sapphire] (20, 1) rectangle (40, 3);

    \draw[draw=text, fill=pink] (0, 9) rectangle (16, 11);

    \draw[draw=text, fill=pink] (16, 7) rectangle (32, 9);


    \draw[draw=text, fill=pink] (32, 11) rectangle (40, 13);

    \draw[draw=text, fill=rosewater] (0, 15) rectangle (24, 17);
    
    \draw[draw=text, fill=rosewater] (24, 13) rectangle (40, 15);


    \draw[dashed, draw=text] (8,0) -- ++(0,18);
    \draw[dashed, draw=text] (16,0) -- ++(0,18);
    \draw[dashed, draw=text] (24,0) -- ++(0,18);
    \draw[dashed, draw=text] (32,0) -- ++(0,18);
    \draw[dashed, draw=text] (40,0) -- ++(0,18);
    
\end{scope}

\draw[fill=base] (2.75,-5.5) rectangle ++(2.25,1) node[midway, align=center] {Shift Plan \& \\ Human Factors};

\draw[-latex] (1.125, -4.5) -- ++(0, -1.75);
\draw[-latex] (3.875, -5.5) -- ++(0, -0.75);

\draw[fill=base] (0,-7.75) rectangle ++(5,1.5) node[midway, align=center] {Worker–Line Allocation \\ (Markov Decision Process)};

\draw[-latex] (2.5, -7.75) -- ++(0, -0.75);

\begin{scope}[shift={(0,-11)}, scale=0.125,]
    \draw[-latex] (0,19) node[above] {} -- (0,0) --  (44,0);

    \draw[ultra thick, draw=rosewater] (0,-9) -- ++(4,0) node[right]{Line $l_1$};
    \draw[ultra thick, draw=pink] (16,-9) -- ++(4,0) node[right]{Line $l_2$};
    \draw[ultra thick, draw=sapphire] (32,-9) -- ++(4,0) node[right]{Line $l_3$};

    \foreach \x/\xtext in {4, 12,...,40}
    \draw[shift={(\x,0)}] (0,0) -- (0,-0.75);
    
    \foreach \x/\xtext in {0, 8,...,40}
    \draw[shift={(\x,0)}] (0,0) -- (0,-1);

    \node[align=left, anchor=west] at (-10, 16) {\tiny Geometry 8};
    \node[align=left, anchor=west] at (-10, 14) {\tiny Geometry 7};
    \node[align=left, anchor=west] at (-10, 12) {\tiny Geometry 6};
    \node[align=left, anchor=west] at (-10, 10) {\tiny Geometry 5};
    \node[align=left, anchor=west] at (-10, 8) {\tiny Geometry 4};
    \node[align=left, anchor=west] at (-10, 6) {\tiny Geometry 3};
    \node[align=left, anchor=west] at (-10, 4) {\tiny Geometry 2};
    \node[align=left, anchor=west] at (-10, 2) {\tiny Geometry 1};
    
    \node[align=left, anchor=west] at (45,0) {\scriptsize time};
    \node[align=center, anchor=north, yshift=0.1mm] at (48.5,-1) {\tiny [min]};
    \node[align=center, anchor=north, yshift=0.1mm] at (48.5,-3) {\tiny [Day hh:mm]};
    
    \node[align=center, anchor=north] at (0,-1) {\tiny 0};
    \node[align=center, anchor=north] at (8,-1) {\tiny 480};
    \node[align=center, anchor=north] at (16,-1) {\tiny 960};
    \node[align=center, anchor=north] at (24,-1) {\tiny 1440};
    \node[align=center, anchor=north] at (32,-1) {\tiny 1920};
    \node[align=center, anchor=north] at (40,-1) {\tiny 2400};

    \node[align=center, anchor=north] at (0,-3) {\tiny Mon 6:00};
    \node[align=center, anchor=north] at (8,-3) {\tiny Mon 14:00};
    \node[align=center, anchor=north] at (16,-3) {\tiny Mon 22:00};
    \node[align=center, anchor=north] at (24,-3) {\tiny Tue 6:00};
    \node[align=center, anchor=north] at (32,-3) {\tiny Tue 14:00};
    \node[align=center, anchor=north] at (40,-3) {\tiny Tue 22:00};

    \draw[draw=text, fill=sapphire] (0, 3) rectangle (8, 5);
    \filldraw[fill=chart-red, draw=text] (1, 4) circle [radius=0.4]; 
    \filldraw[fill=chart-green, draw=text] (2, 4) circle [radius=0.4]; %
    \filldraw[fill=chart-blue, draw=text] (3, 4) circle [radius=0.4]; %
    \filldraw[fill=chart-orange, draw=text] (4, 4) circle [radius=0.4]; %

    \draw[draw=text, fill=sapphire] (8, 5) rectangle (20, 7);
    \filldraw[fill=css-AliceBlue, draw=text] (9, 6) circle [radius=0.4]; %
    \filldraw[fill=css-AntiqueWhite, draw=text] (10, 6) circle [radius=0.4]; %
    \filldraw[fill=css-Aqua, draw=text] (11, 6) circle [radius=0.4]; %

    \filldraw[fill=css-Chartreuse, draw=text] (17, 6) circle [radius=0.4];
    \filldraw[fill=css-Chocolate, draw=text] (18, 6) circle [radius=0.4];
    \filldraw[fill=css-Coral, draw=text] (19, 6) circle [radius=0.4];
   
    \draw[draw=text, fill=sapphire] (20, 1) rectangle (40, 3);
    \filldraw[fill=css-Chartreuse, draw=text] (21, 2) circle [radius=0.4];
    \filldraw[fill=css-Chocolate, draw=text] (22, 2) circle [radius=0.4];
    \filldraw[fill=css-Coral, draw=text] (23, 2) circle [radius=0.4];
    \filldraw[fill=chart-maroon, draw=text] (25, 2) circle [radius=0.4];
    \filldraw[fill=chart-beige, draw=text] (26, 2) circle [radius=0.4];
    \filldraw[fill=chart-teal, draw=text] (27, 2) circle [radius=0.4];
    \filldraw[fill=css-BlanchedAlmond, draw=text] (33, 2) circle [radius=0.4];
    \filldraw[fill=css-BurlyWood, draw=text] (34, 2) circle [radius=0.4];
    \filldraw[fill=css-CadetBlue, draw=text] (35, 2) circle [radius=0.4];

    \draw[draw=text, fill=pink] (0, 9) rectangle (16, 11);
    \filldraw[fill=chart-purple, draw=text] (1, 10) circle [radius=0.4]; %
    \filldraw[fill=chart-cyan, draw=text] (2, 10) circle [radius=0.4]; %
    \filldraw[fill=chart-magenta, draw=text] (3, 10) circle [radius=0.4];
    \filldraw[fill=chart-lime, draw=text] (4, 10) circle [radius=0.4]; %
    \filldraw[fill=chart-pink, draw=text] (5, 10) circle [radius=0.4]; %
    \filldraw[fill=css-Aquamarine, draw=text] (9, 10) circle [radius=0.4];%
    \filldraw[fill=css-Azure, draw=text] (10, 10) circle [radius=0.4]; %
    \filldraw[fill=css-Beige, draw=text] (11, 10) circle [radius=0.4];
    \filldraw[fill=css-Bisque, draw=text] (12, 10) circle [radius=0.4]; %
    \filldraw[fill=css-Black, draw=text] (13, 10) circle [radius=0.4]; %

    \draw[draw=text, fill=pink] (16, 7) rectangle (32, 9);
    \filldraw[fill=css-CornflowerBlue, draw=text] (17, 8) circle [radius=0.4];
    \filldraw[fill=css-Cornsilk, draw=text] (18, 8) circle [radius=0.4];
    \filldraw[fill=css-Crimson, draw=text] (19, 8) circle [radius=0.4];
    \filldraw[fill=css-DarkBlue, draw=text] (20, 8) circle [radius=0.4];
    \filldraw[fill=css-DarkCyan, draw=text] (21, 8) circle [radius=0.4];

    \filldraw[fill=chart-green, draw=text] (25, 8) circle [radius=0.4];
    \filldraw[fill=chart-cyan, draw=text] (26, 8) circle [radius=0.4];
    \filldraw[fill=chart-purple, draw=text] (27, 8) circle [radius=0.4];
    \filldraw[fill=chart-mint, draw=text] (28, 8) circle [radius=0.4];
    \filldraw[fill=chart-red, draw=text] (29, 8) circle [radius=0.4]; 

    \draw[draw=text, fill=pink] (32, 11) rectangle (40, 13);
    \filldraw[fill=css-AntiqueWhite, draw=text] (33, 12) circle [radius=0.4];
    \filldraw[fill=css-Brown, draw=text] (34, 12) circle [radius=0.4];
    \filldraw[fill=css-Azure, draw=text] (35, 12) circle [radius=0.4];
    \filldraw[fill=css-Aquamarine, draw=text] (36, 12) circle [radius=0.4];
    \filldraw[fill=css-Blue, draw=text] (37, 12) circle [radius=0.4];
    \filldraw[fill=css-Aqua, draw=text] (38, 12) circle [radius=0.4];

    \draw[draw=text, fill=rosewater] (0, 15) rectangle (24, 17);
    \filldraw[fill=chart-teal, draw=text] (1, 16) circle [radius=0.4]; %
    \filldraw[fill=chart-lavender, draw=text] (2, 16) circle [radius=0.4];
    \filldraw[fill=chart-brown, draw=text] (3, 16) circle [radius=0.4]; %
    \filldraw[fill=chart-beige, draw=text] (4, 16) circle [radius=0.4]; %
    \filldraw[fill=chart-maroon, draw=text] (5, 16) circle [radius=0.4]; %
    \filldraw[fill=chart-mint, draw=text] (6, 16) circle [radius=0.4]; %
    \filldraw[fill=css-BlanchedAlmond, draw=text] (9, 16) circle [radius=0.4]; %
    \filldraw[fill=css-Blue, draw=text] (10, 16) circle [radius=0.4];
    \filldraw[fill=css-BlueViolet, draw=text] (11, 16) circle [radius=0.4]; %
    \filldraw[fill=css-Brown, draw=text] (12, 16) circle [radius=0.4]; %
    \filldraw[fill=css-BurlyWood, draw=text] (13, 16) circle [radius=0.4]; %
    \filldraw[fill=css-CadetBlue, draw=text] (14, 16) circle [radius=0.4]; %
    \filldraw[fill=css-DarkGoldenRod, draw=text] (17, 16) circle [radius=0.4];
    \filldraw[fill=css-DarkGreen, draw=text] (18, 16) circle [radius=0.4];
    \filldraw[fill=css-DarkKhaki, draw=text] (19, 16) circle [radius=0.4];
    \filldraw[fill=css-DarkMagenta, draw=text] (20, 16) circle [radius=0.4];
    \filldraw[fill=css-DarkOliveGreen, draw=text] (21, 16) circle [radius=0.4];
    \filldraw[fill=css-DarkOrange, draw=text] (22, 16) circle [radius=0.4];
    
    \draw[draw=text, fill=rosewater] (24, 13) rectangle (40, 15);
    \filldraw[fill=chart-orange, draw=text] (25, 14) circle [radius=0.4];
    \filldraw[fill=chart-brown, draw=text] (26, 14) circle [radius=0.4];
    \filldraw[fill=chart-blue, draw=text] (27, 14) circle [radius=0.4];
    \filldraw[fill=chart-pink, draw=text] (28, 14) circle [radius=0.4];
    \filldraw[fill=chart-lime, draw=text] (29, 14) circle [radius=0.4];

    \filldraw[fill=css-Black, draw=text] (33, 14) circle [radius=0.4];
    \filldraw[fill=css-BlueViolet, draw=text] (34, 14) circle [radius=0.4];
    \filldraw[fill=css-Bisque, draw=text] (35, 14) circle [radius=0.4];
    \filldraw[fill=css-Beige, draw=text] (36, 14) circle [radius=0.4];
    \filldraw[fill=css-AliceBlue, draw=text] (37, 14) circle [radius=0.4];

    \draw[dashed, draw=text] (8,0) -- ++(0,18);
    \draw[dashed, draw=text] (16,0) -- ++(0,18);
    \draw[dashed, draw=text] (24,0) -- ++(0,18);
    \draw[dashed, draw=text] (32,0) -- ++(0,18);
    \draw[dashed, draw=text] (40,0) -- ++(0,18);
    
\end{scope}

\end{tikzpicture}
}
\caption{
Two-layer optimization framework. 
The first layer uses a CP solver for Order–Line Allocation, producing a schedule of which geometries are produced on which machines and when. 
The second layer uses an MDP-based approach for Worker–Line Allocation, integrating the schedule with shift plans and human factors to assign workers fairly and efficiently.}
\label{fig:main-example}
\end{figure}
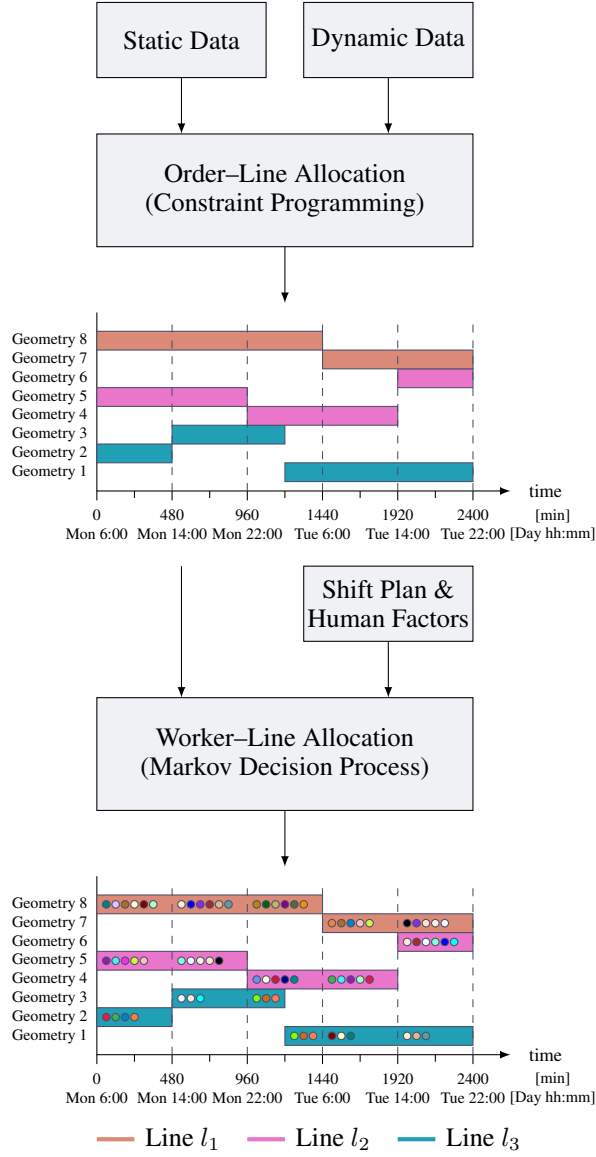

\subsection{Constraint Programming}

CP is a declarative modeling paradigm in which the problem is encoded as variables, domains and constraints. 
A solver searches for assignments that satisfy all constraints and (optionally) optimize an objective function. 
In contrast to procedural encodings, CP emphasizes what must hold rather than how to compute it. 
For scheduling problems, CP provides native modeling primitives such as interval variables, no-overlap constraints, and optional intervals. These can be used to represent deadlines, start and end times, machine assignment indicators, and to formulate both the feasibility constraints and the cost function for optimization.
In our work, CP offered several advantages that aligned well with the requirements of the scheduling subproblem. 
The modeling of temporal constraints such as setup intervals, processing intervals, and alternative machine assignments could be expressed using a small set of high-level constructs, resulting in compact formulations. 
CP’s native temporal primitives, particularly interval variables combined with no-overlap constraints, allowed us to capture machine capacity restrictions and sequencing requirements directly, avoiding the modeling overhead typical of purely linear formulations. 

\subsection{Order–Line Allocation Constraint Programming Definition}
\label{sec:order-line-def}

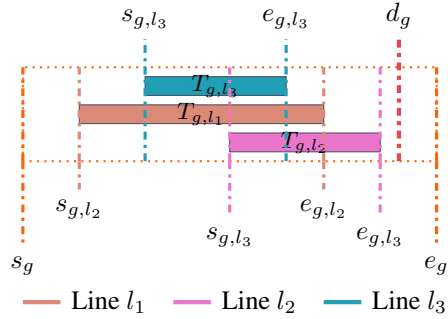
\begin{figure}[htb!]
    \centering
    \begin{tikzpicture}

    \begin{scope}[shift={(0,-11)}, scale=0.125,]


    \draw[draw=text, fill=pink] (22, 0) rectangle ++(16, 2) node[midway] {\footnotesize $T_{g,l_2}$};
    \draw[draw=text, fill=rosewater] (6, 3) rectangle ++(26, 2) node[midway] {\footnotesize $T_{g,l_1}$};
    \draw[draw=text, fill=sapphire] (13, 6) rectangle ++(15, 2) node[midway] {\footnotesize $T_{g,l_3}$};
    
    \draw[draw=peach, thick, dotted] (0, -1)  rectangle ++(44, 10); 
    
    \draw[draw=maroon, ultra thick, dash dot] (40, -1) -- ++(0,10) -- ++(0, 3) node[above] {$d_{g}$};

    \draw[draw=peach, very thick, dash dot] (0, -1) edge ++(0,10) -- ++(0, -9) node[below] {$s_{g}$};
    \draw[draw=peach, very thick, dash dot] (44, -1) edge ++(0,10) -- ++(0, -9)node[below] {$e_{g}$};

    \draw[draw=pink, very thick, dash dot] (22, -1) edge ++(0,10) -- ++(0, -6) node[below] {$s_{g,l_3}$};
    \draw[draw=pink, very thick, dash dot] (38, -1) edge ++(0,10) -- ++(0, -6) node[below] {$e_{g,l_3}$};

    \draw[draw=rosewater, very thick, dash dot] (6, -1) edge ++(0,10)  -- ++(0, -3) node[below] {$s_{g, l_2}$};
    \draw[draw=rosewater, very thick, dash dot] (32, -1) edge ++(0,10)  -- ++(0, -3)  node[below] {$e_{g, l_2}$};

    \draw[draw=sapphire, very thick, dash dot] (13, -1) -- ++(0,10) -- ++(0, 3) node[above] {$s_{g, l_3}$};
    \draw[draw=sapphire, very thick, dash dot] (28, -1) -- ++(0,10) -- ++(0, 3) node[above] {$e_{g, l_3}$};

    \draw[ultra thick, draw=rosewater] (0,-16) -- ++(4,0) node[right]{Line $l_1$};
    \draw[ultra thick, draw=pink] (16,-16) -- ++(4,0) node[right]{Line $l_2$};
    \draw[ultra thick, draw=sapphire] (32,-16) -- ++(4,0) node[right]{Line $l_3$};

    \end{scope}
        
    \end{tikzpicture}
\caption{
Decision variables introduced for a batch of geometries $g$ in the Order–Line Allocation model. 
Each geometry may be produced on multiple lines, with production time depending on throughput and setup times.
}
    \label{fig:g-vars-alternatives}
\end{figure}

\begin{figure}[htb!]
    \centering
    \begin{tikzpicture}

    \begin{scope}[shift={(0,0)}, scale=0.125,]
    \draw[-latex] (0,37) node[above] {} -- (0,0) --  (44,0);

    \foreach \x/\xtext in {4, 12,...,40}
    \draw[shift={(\x,0)}] (0,0) -- (0,-0.75);
    
    \foreach \x/\xtext in {0, 8,...,40}
    \draw[shift={(\x,0)}] (0,0) -- (0,-1);

    \node[align=left, anchor=west] at (-10, 33.5) {\tiny Geometry 8};
    \node[align=left, anchor=west] at (-10, 26.5) {\tiny Geometry 5};
    \node[align=left, anchor=west] at (-10, 19.5) {\tiny Geometry 4};
    \node[align=left, anchor=west] at (-10, 12.5) {\tiny Geometry 3};
    \node[align=left, anchor=west] at (-10, 5.5) {\tiny Geometry 2};
    
    \node[align=left, anchor=west] at (45,0) {\scriptsize time};
    \node[align=center, anchor=north, yshift=0.1mm] at (48.5,-1) {\tiny [min]};
    \node[align=center, anchor=north, yshift=0.1mm] at (48.5,-3) {\tiny [Day hh:mm]};
    
    \node[align=center, anchor=north] at (0,-1) {\tiny 0};
    \node[align=center, anchor=north] at (8,-1) {\tiny 480};
    \node[align=center, anchor=north] at (16,-1) {\tiny 960};
    \node[align=center, anchor=north] at (24,-1) {\tiny 1440};
    \node[align=center, anchor=north] at (32,-1) {\tiny 1920};
    \node[align=center, anchor=north] at (40,-1) {\tiny 2400};

    \node[align=center, anchor=north] at (0,-3) {\tiny Mon 6:00};
    \node[align=center, anchor=north] at (8,-3) {\tiny Mon 14:00};
    \node[align=center, anchor=north] at (16,-3) {\tiny Mon 22:00};
    \node[align=center, anchor=north] at (24,-3) {\tiny Tue 6:00};
    \node[align=center, anchor=north] at (32,-3) {\tiny Tue 14:00};
    \node[align=center, anchor=north] at (40,-3) {\tiny Tue 22:00};


    \draw[draw=text, fill=sapphire] (6, 3) rectangle ++(8, 2);
    \draw[draw=text, fill=rosewater] (5, 6) rectangle ++(10, 2);
    \draw[draw=peach, dotted, thick] (0, 3)  rectangle ++(36, 5); 
    \draw[draw=maroon, thick, dash dot] (9,3) -- ++(0,5);
    
    \draw[draw=text, fill=sapphire] (12, 10) rectangle ++(12, 2);
    \draw[draw=text, fill=pink] (4, 13) rectangle ++(18, 2);
    \draw[draw=peach, thick, dotted] (0, 10)  rectangle ++(36, 5); 
    \draw[draw=red, thick, dash dot] (32,10) -- ++(0,5);

    \draw[draw=text, fill=pink] (15, 17) rectangle ++(16, 2);
    \draw[draw=text, fill=rosewater] (6, 20) rectangle ++(26, 2);
    \draw[draw=peach, thick, dotted] (0, 17)  rectangle ++(36, 5); 
    \draw[draw=maroon, thick, dash dot] (14,17) -- ++(0,5);
    
    \draw[draw=text, fill=pink] (13, 24) rectangle ++(16, 2);
    \draw[draw=text, fill=sapphire] (11, 27) rectangle ++(22, 2);
    \draw[draw=peach, thick, dotted] (0, 24)  rectangle ++(36, 5); 
    \draw[draw=maroon, thick, dash dot] (34,24) -- ++(0,5);

    \draw[draw=text, fill=rosewater] (4, 31) rectangle ++(24, 2);
    \draw[draw=text, fill=sapphire] (2, 34) rectangle ++(21, 2);
    \draw[draw=peach, thick, dotted] (0, 31)  rectangle ++(36, 5); 
    \draw[draw=maroon, thick, dash dot] (8, 31) -- ++(0,5);

    \end{scope}

    \draw[-latex] (3.0, -1) -- ++(0, -0.75);

    \begin{scope}[shift={(0,-4.25)}, scale=0.125,]
    \draw[-latex] (0,19) node[above] {} -- (0,0) --  (44,0);

    \draw[ultra thick, draw=rosewater] (0,-9) -- ++(4,0) node[right]{Line $l_1$};
    \draw[ultra thick, draw=pink] (16,-9) -- ++(4,0) node[right]{Line $l_2$};
    \draw[ultra thick, draw=sapphire] (32,-9) -- ++(4,0) node[right]{Line $l_3$};

    \foreach \x/\xtext in {4, 12,...,40}
    \draw[shift={(\x,0)}] (0,0) -- (0,-0.75);
    
    \foreach \x/\xtext in {0, 8,...,40}
    \draw[shift={(\x,0)}] (0,0) -- (0,-1);

    \node[align=left, anchor=west] at (-10, 15) {\tiny Geometry 8};
    \node[align=left, anchor=west] at (-10, 12) {\tiny Geometry 5};
    \node[align=left, anchor=west] at (-10, 9) {\tiny Geometry 4};
    \node[align=left, anchor=west] at (-10, 6) {\tiny Geometry 3};
    \node[align=left, anchor=west] at (-10, 3) {\tiny Geometry 2};

    \node[align=left, anchor=west] at (45,0) {\scriptsize time};
    \node[align=center, anchor=north, yshift=0.1mm] at (48.5,-1) {\tiny [min]};
    \node[align=center, anchor=north, yshift=0.1mm] at (48.5,-3) {\tiny [Day hh:mm]};
    
    \node[align=center, anchor=north] at (0,-1) {\tiny 0};
    \node[align=center, anchor=north] at (8,-1) {\tiny 480};
    \node[align=center, anchor=north] at (16,-1) {\tiny 960};
    \node[align=center, anchor=north] at (24,-1) {\tiny 1440};
    \node[align=center, anchor=north] at (32,-1) {\tiny 1920};
    \node[align=center, anchor=north] at (40,-1) {\tiny 2400};

    \node[align=center, anchor=north] at (0,-3) {\tiny Mon 6:00};
    \node[align=center, anchor=north] at (8,-3) {\tiny Mon 14:00};
    \node[align=center, anchor=north] at (16,-3) {\tiny Mon 22:00};
    \node[align=center, anchor=north] at (24,-3) {\tiny Tue 6:00};
    \node[align=center, anchor=north] at (32,-3) {\tiny Tue 14:00};
    \node[align=center, anchor=north] at (40,-3) {\tiny Tue 22:00};

    \draw[draw=text, fill=sapphire] (0, 2) rectangle ++(8, 2);
    \draw[draw=text, fill=sapphire] (8, 5) rectangle ++(12, 2);

    \draw[draw=text, fill=pink] (0, 11) rectangle ++(16, 2);
    \draw[draw=text, fill=pink] (16, 8) rectangle ++(16, 2);

    \draw[draw=text, fill=rosewater] (0, 14) rectangle ++(24, 2);

\end{scope}
        
    \end{tikzpicture}
\caption{
Illustration of a CP solution. 
For each geometry, one alternative from the feasible set is chosen such that the resulting rectangles do not overlap on the time axis.
}
    \label{fig:cp-example}
\end{figure}
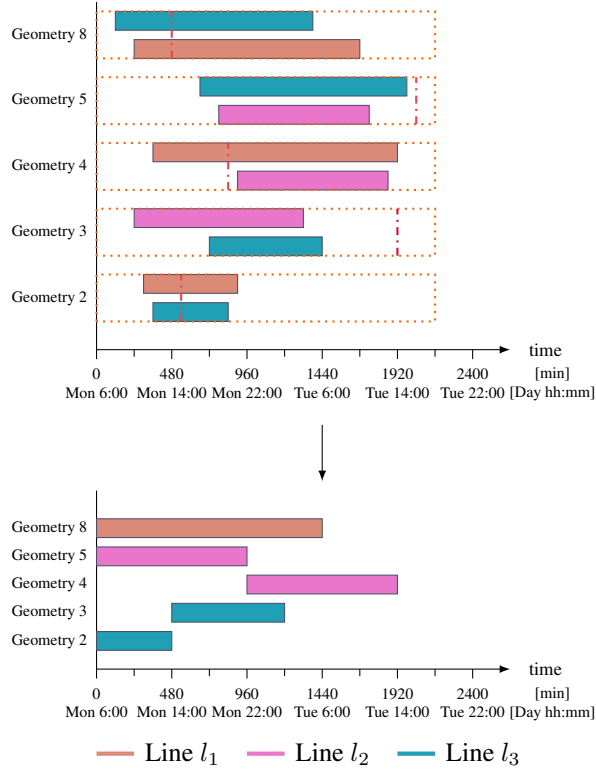

The first-layer addresses the assignment of geometries (products) to lines (machines) and their scheduling over time. 
Let $\mathcal{G}$ denote the set of geometries that have been ordered by customers, each associated with a specified quantity and deadline.
Let $\mathcal{L}$ denote the set of available lines. 
Note that the same geometry may appear multiple times in $\mathcal{G}$ with different quantities and deadlines, as it may belong to different orders. 
For each geometry $g \in \mathcal{G}$, let $A_g \subseteq \mathcal{L}$ denote the set of admissible lines on which $g$ can be processed, determined by technical compatibility. 
The variables defined for each geometry are summarized in Table \ref{tab:g-vars}.
\begin{table*}[htb!]
\centering
\caption{Decision variables for geometry scheduling in the first-layer CP model.}
\label{tab:g-vars}

\begin{tabular*}{\textwidth}{@{\extracolsep\fill}lccp{9cm}@{\extracolsep\fill}}
\toprule
\textbf{Name} & \textbf{Symbol} & \textbf{Type} & \textbf{Description} \\
\midrule

Production start time & $s_{g}$ & Integer & Time at which production of the batch of geometry $g$ begins. \\

Production end time & $e_{g}$ & Integer & Time at which production of the batch of geometry $g$ completes. \\

Production interval & $T_g$ & Interval & Time interval representing the duration of production for batch $g$, defined by $s_g$ and $e_g$. \\

Line start time & $s_{g,l}$ & Integer & Time at which production of batch $g$ begins on line $l$. \\

Line end time & $e_{g,l}$ & Integer & Time at which production of batch $g$ ends on line $l$, including setup and processing times. \\

Line interval & $T_{g,l}$ & Interval & Time interval representing the production of batch $g$ on line $l$, defined by $s_{g,l}$ and $e_{g,l}$. \\

Line selected & $\gamma_{g,l}$ & Boolean & Indicated if the Intervall $T_{g,l}$ is selected in the resulting schedule. \\

Due Date & $d_g$ & Integer & Latest allowable completion time for batch $g$ to ensure on-time delivery to the customer. \\

Priority & $p_g$ & Boolean & Indicates whether batch $g$ has high priority ($p_g = 1$) and must be scheduled before non-priority batches ($p_g = 0$). \\

\bottomrule
\end{tabular*}
\end{table*}
Each geometry $g$ has a due date $d_g$ and a priority flag $p_g \in \{0,1\}$, where $P_g = 1$ indicates a priority geometry that must be scheduled before non-priority geometries. 
A batch of geometry $g$ can be produced on any line $l \subseteq A_g$. 
Each line has distinct characteristics, resulting in different setup and processing times depending on the assigned line. 
Consequently, the intervals $I_{g,l}$ for producing $g$ may have different durations. 
Figure \ref{fig:g-vars-alternatives} illustrates an example with three admissible lines, highlighting this behavior.
In Figure \ref{fig:g-vars-alternatives}, three admissible lines $l_1$, $l_2$, and $l_3$ for producing geometry $g$ are visualized. 
The intervals $T_{g,l_1}$, $T_{g,l_2}$, and $T_{g,l_3}$ have different lengths, reflecting variations in setup and processing times. 
For each geometry $g$, exactly one interval $T_{g,l}$ must be selected. 
When an interval is selected, the corresponding global interval $T_g$ is enforced. 
The global interval $T_g$, along with $s_g$ and $e_g$, is subsequently used to enforce priority constraints and to formulate the objective function. 
The scheduling problem can be visualized as selecting one interval $T_{g,l}$ for each geometry $g$ and arranging these intervals in time such that no intervals assigned to the same line overlap. 
The selected intervals must be arranged to minimize makespan and tardiness, as illustrated in Figure \ref{fig:cp-example}.
The possible lines for a batch of geometries are indicated by an orange dotted rectangel.
Each of these rectangles introduces decision variabls as indicated by Figure \ref{fig:g-vars-alternatives} and Table \ref{tab:g-vars}.
On a high level a CP-solver systematically tries out different cominations of line intervals and shifts the intervals around on the x-axis, such that the objective function is optimized. 

The dynamics of the scheduling problem can be formally expressed using mathematical equations.
The objective function $Z$, which is to be minimized, is defined as:
\begin{align}
Z &= w_c \cdot C_{\max} + w_{\tau} \cdot \sum_{g \in \mathcal{G}} \tau_g  && w_c \geq 0, \;  w_{\tau} \geq 0 \label{eq:objective-def} \\
\tau_g &= \max(0, e_g - d_g) &&\forall \: g \in \mathcal{G} \label{eq:tardiness-def} \\
C_{\max} &= \max_{g \in \mathcal{G}} e_g &&\forall \: g \in \mathcal{G} \label{eq:makespan-def}
\end{align}
The objective function $Z$ is a weighted sum of two performance metrics: the total tardiness $\tau_g$ and the makespan $C_{\max}$.  
Tardiness quantifies the delay of each job $g \in \mathcal{G}$ relative to its due date $d_g$, and is set to zero whenever the job finishes on time or earlier.  
The makespan $C_{\max}$ measures the overall length of the schedule, i.e., the completion time of the last job in the system.  
Minimizing tardiness encourages the schedule to align job completion times closely with their due dates, which often leads to prioritization of time-critical jobs.  
In contrast, minimizing makespan seeks to compress the entire schedule and maximize throughput, which may conflict with tardiness minimization.
The weights $w_c$ and $w_{\tau}$ act as hyperparameters of the first-layer, controlling the trade-off between throughput optimization and tardiness minimization.  
By adjusting these weights, decision-makers can emphasize either makespan reduction or tardiness minimization depending on production priorities.  
The constraints of the scheduling problem in the first layer can be formalized by the following set of equations:
\begin{align}
s_g &= \sum_{l \in A_g} x_{g,l} \cdot s_{g,l}  &&\forall \: g \in \mathcal{G} &&  \label{eq:start-time-selection} \\
e_g &= \sum_{l \in A_g} x_{g,l} \cdot e_{g,l}  &&\forall \: g \in \mathcal{G} && \label{eq:end-time-selection} \\
T_{g,l} &= [s_{g,l}, e_{g,l}] &&\forall \: g \in \mathcal{G}, &&\forall \: l \in A_g \label{eq:interval-definition} \\
s_{g,l} &\ge e_{h,l} \; \lor \; s_{h,l} \ge e_{g,l}  &&\forall \: g,h \in \mathcal{G}, &&g \neq h, \; \forall \: l \in \mathcal{L} \label{eq:no-overlap} \\
e_g &\le s_h  &&\forall \: g,h \in \mathcal{G}, && p_g = 1, \; p_h = 0 \label{eq:priority-constraint} \\
1 &= \sum_{l \in A_g} \gamma_{g,l}  &&\forall \: g \in \mathcal{G},  && \gamma_{g,l} \in \{0,1\} \label{eq:line-selection}
\end{align}
The above constraints ensure that the schedule is both feasible and consistent with the production requirements.  
Equations \eqref{eq:start-time-selection} and \eqref{eq:end-time-selection} link the global start and end times $s_g$ and $e_g$ of each geometry $g$ to the start and end times on the selected line using the binary decision variables $\gamma_{g,l}$.  
Equation \eqref{eq:interval-definition} formally defines each line interval $T_{g,l}$, ensuring it is characterized by its start and end times.  
The no-overlap constraint in \eqref{eq:no-overlap} enforces that two geometries $g$ and $h$ assigned to the same line $l$ do not overlap in time, thereby respecting machine capacity limitations.  
The priority constraint \eqref{eq:priority-constraint} guarantees that high-priority geometries ($p_g = 1$) are completed before any non-priority geometry ($p_h = 0$) begins, supporting due-date adherence for urgent orders.  
Finally, the line-selection constraint \eqref{eq:line-selection} ensures that exactly one admissible line is chosen for the production of each geometry.  
Together, these constraints form the core of the first-layer CP model, defining a combinatorial search space in which the solver systematically explores alternative line assignments and production sequences to optimize the weighted objective function \eqref{eq:objective-def}.

It is worth noting that this formulation constitutes a variation of the classical Flexible Job-Shop Problem (FJSP).  
In a standard FJSP, each job typically consists of a sequence of operations with strict technological precedence constraints, and the problem involves both selecting a machine for each operation and sequencing operations across machines.  
In contrast, our formulation considers each job as a single operation, a batch of geometries, and therefore omits operation-precedence constraints altogether.  
This reduction simplifies the problem structure and enables the CP approach to be applied efficiently to this still NP-hard problem on industrial time scales.  

\subsection{Markov Decision Process}

Many sequential decision-making problems can be formally described as a \textit{Markov Decision Process} (MDP).  
An MDP provides a mathematical framework for modeling decision problems where outcomes are partly under the control of a decision-maker and partly determined by the environment.  
Formally, an MDP is defined by the tuple $\langle \mathcal{S}, \mathcal{A}, P, R, \gamma \rangle$, where:

\begin{itemize}
    \item $\mathcal{S}$ is the set of all possible states that describe the environment at a given time step.
    \item $\mathcal{A}$ is the set of all actions available to the decision-maker.
    \item $P(s' \mid s, a)$ defines the transition probability of reaching state $s'$ after taking action $a$ in state $s$.
    \item $R(s,a)$ is the reward function, returning a scalar value that quantifies the desirability of taking action $a$ in state $s$.
    \item $\Gamma \in [0,1]$ is a discount factor that determines the relative importance of future rewards compared to immediate ones.
\end{itemize}

A key property of an MDP is the \textit{Markov property}, which states that the next state $s'$ depends only on the current state $s$ and the action $a$, and not on the sequence of previous states or actions.  
This property allows compact modeling of complex systems and enables the use of efficient algorithms for planning and learning.

The goal in an MDP is to find \textit{policy} $\Pi(a \mid s)$, which is a mapping from states to actions (deterministically or stochastically), that maximizes the expected cumulative reward, also called the \textit{return}.  
This is typically formalized as:
\begin{equation}
J(\Pi) = \mathbb{E}_{\Pi}\!\left[\sum_{t=0}^{\infty} \Gamma^t R(s_t, a_t)\right]
\end{equation}
where $s_t$ and $a_t$ denote the state and action at time step $t$, respectively.

A wide range of algorithms exist for solving MDPs.  
\textit{Model-free} approaches, such as Q-learning or policy gradient methods, learn directly from experience without requiring knowledge of the transition probabilities $P$.  
\textit{Model-based} approaches, in contrast, rely on an explicit or learned model of $P$ to plan ahead by simulating the consequences of potential actions before taking them in the real environment.  
Hybrid methods, such as actor–critic algorithms, combine value-based and policy-based learning to leverage the strengths of both.

MDPs can also serve as the foundation for more advanced planning approaches, such as neural MCTS.  
MCTS is a heuristic search method originally developed for combinatorial games, in which an agent explores possible future sequences of actions to estimate which action will likely lead to the most favorable outcome.  
Neural MCTS combines this look-ahead planning with neural network approximations of policy and value functions, allowing the agent to generalize across similar states \citep{kemmerling2024beyond}.  

While both RL and neural MCTS aim to produce policies and value estimates, there are important distinctions.  
Traditional RL learns policies from accumulated experience, aiming to generalize to unseen states, and executes actions based on this learned policy without forward planning at decision time.  
In contrast, neural MCTS constructs a policy for each encountered state via multi-step forward search, evaluating hypothetical future trajectories; the computed action is specific to the current state and is not reused for other states \citep{kemmerling2024beyond}.  
Consequently, RL requires potentially expensive training upfront but incurs low computational cost during execution, whereas neural MCTS performs expensive planning at decision time but can leverage model knowledge directly.  

MDPs provide a formal, flexible framework that supports both these approaches, making them particularly suitable for complex, sequential decision-making tasks in operations research, robotics, and production scheduling.

\subsection{Worker–Line Allocation Markov Decision Process Definition}
\label{sec:worker-line-def}

The second layer of our optimization approach formalizes the worker-line allocation problem as a Markov Decision Process (MDP). 
It takes the output of the Order–Line allocation from the first layer, along with shift plans and human factor data, to generate a schedule that assigns personnel to lines.  
To perform these allocations, we decompose the Order–Line schedule into atomic shop floor situations.  
Production tasks may span multiple shifts, causing the operating personnel on a line to change at the end of a shift.  
Conversely, a task may finish in the middle of a shift, requiring personnel to start working on a new batch of geometries.  
To handle both scenarios, we divide the schedule into discrete time intervals $\delta_n$, which can be interpreted as slots on each line that need to be filled with personnel.  
Within each time interval, the set of available slots and personnel is unique, defining an atomic shop floor situation.  
To construct these time intervals, we first collect all start and end times of tasks, along with the start and end times of shifts. 
This set is then sorted in ascending order, and consecutive elements define the boundaries of the intervals $\delta_n$.  
Figure \ref{fig:how-to-time-inervals} illustrates the resulting intervals for the example introduced in Figure \ref{fig:cp-example}.  
Interval boundaries are marked at shift changes (e.g., 6:00, 14:00, 22:00), while some intervals, such as $\delta_3$, end in the middle of a shift due to task changes on line $l_3$.  
Empty white circles represent slots that require assignment of personnel.  
This discretization into intervals is crucial for defining the \textit{state space}, \textit{action space}, and \textit{reward function} of the MDP.  
It allows the problem to be modeled as a sequence of decisions, where the allocation in each interval influences both future personnel availability and fairness outcomes.  
By formulating the problem as an MDP, we can systematically account for worker preferences, experience, medical constraints, and task resilience while generating schedules that balance productivity and fairness.  

The following subsections detail the design of the core MDP components: the observation space, action space, and reward function.  
The dynamics of the assignment problem are discussed together with the observation space, as the chosen data representation is tightly coupled with the state transitions in this use case.

\subsubsection{State Space}
\label{sec:state-space}
\begin{figure}[htb]
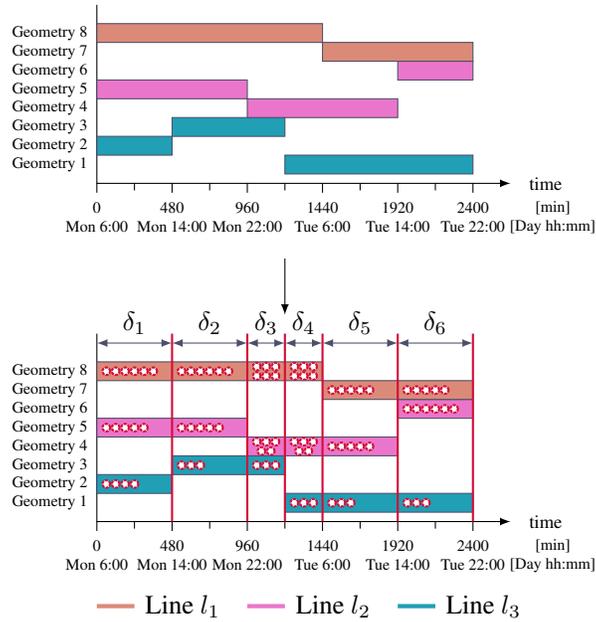

\centerline{

}
\caption{
Construction of time intervals $\delta_n$ from task and shift boundaries. 
White circles indicate personnel slots that must be filled in subsequent Worker–Line Allocation.
}
\label{fig:how-to-time-inervals}
\end{figure}
\begin{figure*}[htb!]
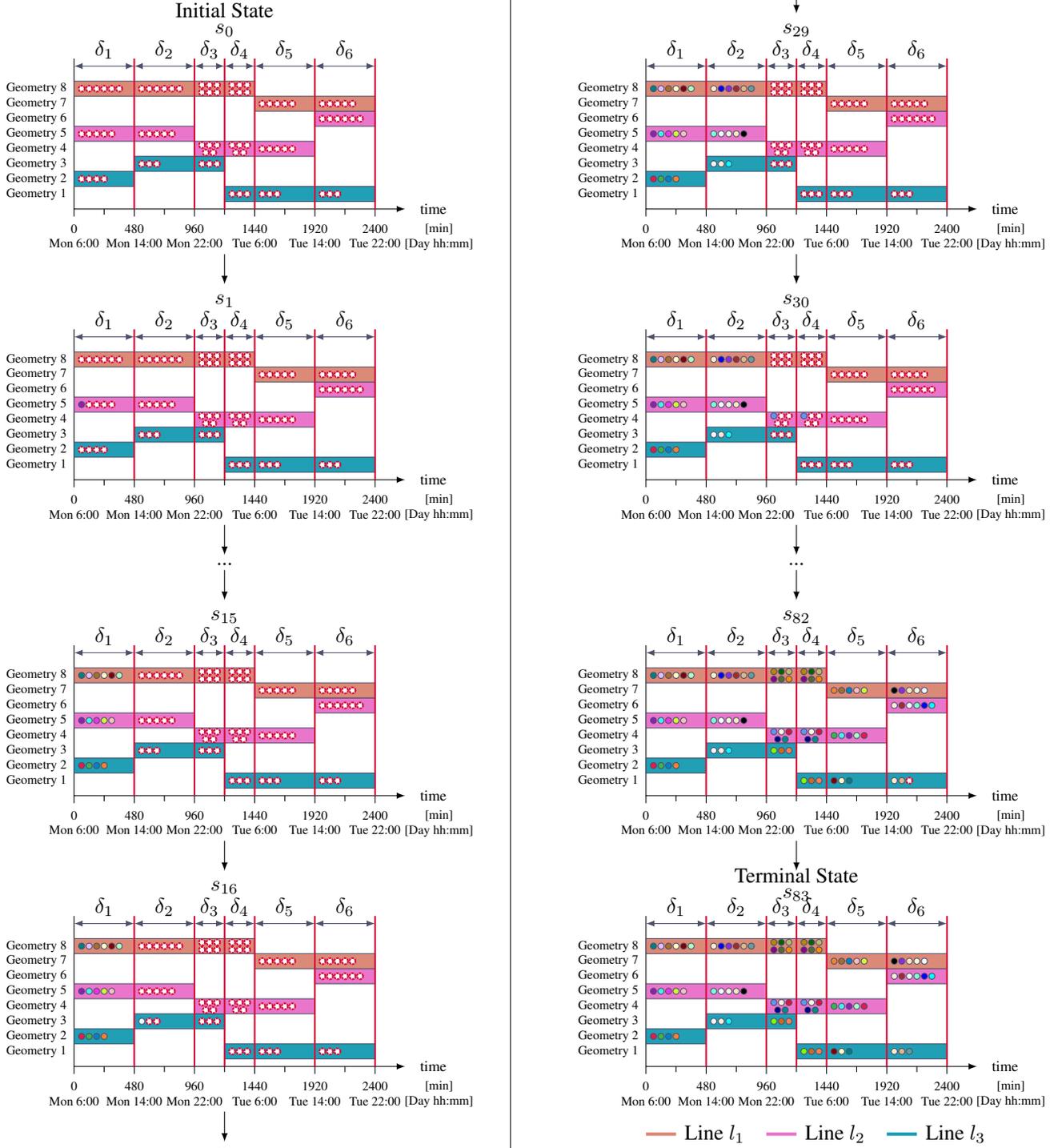

\centerline{

}
\caption{
Example state sequence for solving a Worker-Line Allocation problem, illustrating the evolution of allocations across successive MDP steps. 
Transitions $s_0 \rightarrow s_1$, $s_{15} \rightarrow s_{16}$, $s_{82} \rightarrow s_{29}$, and $s_{30} \rightarrow s_{83}$ highlight the assignment of a worker to an empty slot (white circle with red dashed border), with $s_0$ also showing the progression from $\delta_1$ to $\delta_2$. 
}
\label{fig:mdp-process-example}
\end{figure*}

The state of the MDP is represented as a structured table of numerical values. 
Table~\ref{tab:state-space-table} shows an example state from the solution process of the instance introduced in Figure~\ref{fig:main-example}. 
The state corresponds to $s_{16}$ in the state sequence illustrated in Figure~\ref{fig:mdp-process-example}. 
For each time interval $\delta_n$, one row per line is created, yielding a row for every \textit{line–interval} combination. 
Each row contains all information required for personnel allocation in the given interval and can be conceptually interpreted as a rectangle in Figure~\ref{fig:mdp-process-example}, extended with workforce requirements and human-factor attributes.
\begin{table*}[htb]
    \centering
    \scriptsize
\caption{
Example state representation for a Worker–Line Allocation instance. Each row represents a line in a time interval, including personnel requirements, current allocations, and human factor attributes of all available workers. 
The illustrated state corresponds to state $s_{16}$ in Figure \ref{fig:mdp-process-example}.
}
    \label{tab:state-space-table}
    \begin{tabular*}{\textwidth}{{lllllll|lllllllllllll}}%
        \hline
\multirow{2}{*}{$\delta_n$} & $s_{\delta_n}$ & $e_{\delta_n}$ & \multirow{2}{*}{Line} & \multirow{2}{*}{Req.} & \multirow{2}{*}{Alloc.} & \multirow{2}{*}{Done} & \multicolumn{6}{c}{$\omega_1$}                       & ...  & \multicolumn{6}{c}{$\omega_k$}                       \\
                            & $[min]$          & $[min]$          &                       &                       &                         &                       & $\alpha$ & $\mu$ & $\pi$ & $\rho$ & $\xi$ & $\sigma$ &   & $\alpha$ & $\mu$ & $\pi$ & $\rho$ & $\xi$ & $\sigma$ \\
\hline
1                           & 0              & 480            & $l_1$                 & 6                     & 6                       & 1                     & 1        & 1     & 0.22  & 0.73   & 0.33  & 0        & ... & 0        & 1     & 0.56  & 0.93   & 1.0   & 0        \\
1                           & 0              & 480            & $l_2$                 & 5                     & 5                       & 1                     & 1        & 1     & 0.84  & 0.79   & 0.83  & 1        & ... & 0        & 1     & 0.85  & 0.95   & 0.84  & 0        \\
1                           & 0              & 480            & $l_3$                 & 4                     & 4                       & 1                     & 1        & 0     & 0     & 0      & 0     & 0        & ... & 0        & 1     & 0.60  & 0.49   & 0.51  & 0        \\
2                           & 480            & 960            & $l_1$                 & 6                     & 1                       & 0                     & 0        & 1     & 0.22  & 0.73   & 0.33  & 0        & ... & 1        & 1     & 0.56  & 0.93   & 1.0   & 1        \\
2                           & 480            & 960            & $l_2$                 & 5                     & 0                       & 0                     & 0        & 1     & 0.84  & 0.79   & 0.83  & 0        & ... & 1        & 1     & 0.85  & 0.95   & 0.84  & 0        \\
2                           & 480            & 960            & $l_3$                 & 4                     & 4                       & 1                     & 0        & 0     & 0.66  & 0.39   & 0.31  & 0        & ... & 1        & 1     & 0.24  & 0.56   & 0.93  & 0        \\
3                           & 960            & 1200           & $l_1$                 & 6                     & 0                       & 0                     & 0        & 1     & 0.22  & 0.73   & 0.33  & 0        & ... & 0        & 1     & 0.36  & 0.41   & 0.95  & 0        \\
3                           & 960            & 1200           & $l_2$                 & 5                     & 0                       & 0                     & 0        & 1     & 0.61  & 0.27   & 0.26  & 0        & ... & 0        & 0     & 0     & 0      & 0     & 0        \\
3                           & 960            & 1200           & $l_3$                 & 4                     & 0                       & 0                     & 0        & 1     & 0.66  & 0.39   & 0.31  & 0        & ... & 0        & 1     & 0.24  & 0.56   & 0.93  & 0        \\
...                         & ...            & ...            & ...                   & ...                   & ...                     & ...                   & ...      & ...   & ...   & ...    & ...   & ...      & ... & ...      & ...   & ...   & ...    & ...   & ...     
\end{tabular*}
\end{table*}
Each row contains: the interval identifier $\delta_n$, its start and end times $s_{\delta_n}$, $e_{\delta_n}$, and the line identifier $l$. 
Together, $(l, s_{\delta_n}, e_{\delta_n})$ uniquely define a set of assignment slots. 
The column \textit{Req.} specifies the required number of workers, corresponding to the number of white circles in Figure~\ref{fig:how-to-time-inervals}. 
The column \textit{Alloc.} counts how many workers are already assigned in the current partial solution, while the binary flag \textit{Done} indicates whether further allocations are possible (\textit{Done} = 1 if all slots are filled or no feasible worker remains).
For each worker $\omega$, six attributes are considered:

\begin{itemize}
    \item \textbf{Availability} $\alpha \in \{0,1\}$ – indicates whether the worker is present during the interval $\delta_n$. Values of $\alpha$ can be obtained from the shift plan.
    
    \item \textbf{Medical condition} $\mu \in \{0,1\}$ – equals $1$ if the worker is medically cleared to operate the given line–geometry combination. The values of $\mu$ are available in the static data tables (cf. Figure~\ref{fig:main-example}).
    
    \item \textbf{Preference} $\pi \in [0,1]$ – quantifies the worker’s preference for the assigned task, with $1$ indicating maximum preference. Preferences are collected via questionnaires assessing how much a worker prefers to work on a specific line and geometry combination.
    
    \item \textbf{Resilience} $\rho \in [0,1]$ – represents the physical and cognitive strain of the task, with higher values indicating less demanding work. 
    These values are derived from wearable sensors that record biometric indicators of exertion, which are then aggregated into a single normalized score \citep{paletta2024resilience}.
    
    \item \textbf{Experience} $\xi \in [0,1]$ – captures the worker’s proficiency and training level on the specific line and geometry.
    
    \item \textbf{Allocation flag} $\sigma \in \{0,1\}$ – equals $1$ if the worker is currently assigned to this row in the partial solution.
\end{itemize}

Assessing human resilience—defined as the ability to cope with stress and recover from setbacks—is particularly important in production environments, as it helps prevent stress-related errors, accidents, sick leave, and absenteeism, while promoting well-being, sustainable performance, and efficient worker allocation. 
For quantitative monitoring, wearable sensor data (e.g., heart rate, heart rate variability, temperature) are collected to capture physiological and cognitive-emotional strain. 
These signals are processed and aggregated over a defined time window to represent mental exhaustion \cite{paletta2024resilience}. 
This model can be further enhanced by analyzing recovery dynamics, measured using the psychological construct of the recovery–stress state \citep{kellmann2024recovery, paletta2025recovery}. 
The resulting resilience score is then normalized to a value between 0 and 1, providing an inverse measure of mental exhaustion and reflecting the individual’s adaptive capacity.

The scheduling task consists of selecting, for each row, a subset of workers satisfying $\alpha = 1$ and $\mu = 1$, while maximizing $\pi$, $\rho$, and $\xi$, and distributing preferences as evenly as possible across the workforce to promote fairness. 
Allocation proceeds sequentially across time intervals: workers are first assigned for $\delta_1$, then for $\delta_2$, and so forth. 
When a worker is allocated, $\sigma$ is set to $1$ for the corresponding row and reset to $0$ for all other rows within the same interval, ensuring that each worker is assigned to at most one line per interval. 
The \textit{Alloc.} column is updated accordingly. 
Whenever possible, the MDP environment attempts to allocate the same worker in the subsequent interval on the same line if both the line and geometry remain unchanged, thereby minimizing unnecessary line changes. 
Workers may only switch lines if multiple geometry batches finish simultaneously within a shift, in which case reallocation can improve overall assignment quality and fairness. 
When $\textit{Alloc.} = \textit{Req.}$ or no feasible workers remain, \textit{Done} is set to $1$, enabling the environment to advance even if some slots remain unfilled. 
In practice, unfilled slots trigger human planners to adjust the shift plan or workforce availability before re-running the optimization. 
Once all rows in $\delta_{\hat{n}}$ are marked as done, the allocation process continues with $\delta_{\hat{n}+1}$ until all intervals are processed.
The columns shown in Table~\ref{tab:state-space-table} represent the minimal set required to satisfy the Markov property: past allocations are fully encoded in $\sigma$, and all information required for future decisions is contained within the table, making it a valid MDP state representation. 
The actual implementation additionally tracks setup times, order identifiers, and geometry IDs to facilitate environment dynamics, but these are omitted here for clarity.

Figure~\ref{fig:mdp-process-example} illustrates an example execution of the Worker–Line Allocation process. 
State $s_0$ represents the initial state, where the Order–Line schedule obtained from the first optimization layer is subdivided into time intervals $\delta_n$ as described above. 
Within each interval, several personnel slots (depicted as white circles with red dashed borders) are created for each line. 
At $s_0$, no workers are assigned, so all slots remain empty.
The allocation proceeds sequentially over intervals, beginning with $\delta_1$. 
The currently active interval is highlighted in green with a hatched pattern, previously completed intervals in blue, and future intervals in red. 
In the transition $s_0 \rightarrow s_1$, worker $\omega_1$ (colored purple) is assigned to a previously empty slot for the production of \textit{Geometry~5} on line $l_2$. 
In Table~\ref{tab:state-space-table}, this corresponds to setting $\sigma=1$ for $\omega_1$ in the row representing line $l_2$, while $\sigma=0$ for other rows in the same interval. 
Subsequent assignments in $\delta_1$ proceed analogously until all slots are filled. 
State $s_{15}$ represents the situation immediately after the completion of $\delta_1$: all slots in $\delta_1$ are filled (indicated by blue hatching), and $\delta_2$ is now the active interval (highlighted in green).
In $s_{15} \rightarrow s_{16}$, another worker (colored in \textit{Alice Blue}) is assigned to \textit{Geometry~3} on line $l_3$, and the process continues until all slots in $\delta_2$ are filled.
State $s_{29}$ illustrates the start of allocations for $\delta_3$, which ends mid-shift. 
When a worker is assigned to \textit{Geometry~4} on line $l_2$, the MDP environment also assigns the same worker to line $l_2$ in the subsequent interval $\delta_4$, provided the geometry remains unchanged. 
This mechanism minimizes unnecessary worker line changes and promotes continuity. 
The remaining slots in $\delta_3$ are then filled before proceeding to $\delta_4$, $\delta_5$, and $\delta_6$ in the same manner.
Finally, the transition $s_{82} \rightarrow s_{83}$ represents the last allocation step: a worker is assigned to \textit{Geometry~1} on line $l_3$ in $\delta_6$. 
State $s_{83}$ is the terminal state, characterized by all rows in Table~\ref{tab:state-space-table} having \textit{Done} = 1, indicating that no further allocations are possible.
Table \ref{tab:terminal-state} shows the final schedule in the terminal state.

\begin{table}[htb]
\centering
\caption{
Final schedule with worker assignments produced by the Worker–Line allocation process, corresponding to the terminal state $S_{83}$ in Figure~\ref{fig:mdp-process-example}.
}
\label{tab:terminal-state}
\begin{tabular*}{350pt}{@{\extracolsep\fill}cccccc@{\extracolsep\fill}}%
\toprule
\textbf{Geo.} & \textbf{Start} & \textbf{Finish} & \textbf{Line} & \textbf{Worker Assignments} \\
\midrule
$g_1$ & Tue. 02:00 & Tue. 06:00 & $l_3$ & $\omega_{30}$ $\omega_{31}$ $\omega_{32}$ \\
$g_1$ & Tue. 06:00 & Tue. 14:00 & $l_3$ &  $\omega_{14}$ $\omega_{13}$ $\omega_{10}$\\
$g_1$ & Tue. 14:00 & Tue. 22:00 & $l_3$ &  $\omega_{24}$ $\omega_{28}$ $\omega_{29}$ \\
\midrule
$g_2$ & Mo. 06:00 & Mo. 14:00 & $l_3$ & $\omega_1$ $\omega_{2}$ $\omega_3$ $\omega_{4}$ \\
\midrule
$g_3$ & Mo. 14:00 & Mo. 22:00 & $l_3$ & $\omega_{16}$ $\omega_{17}$ $\omega_{18}$ \\
$g_3$ & Mo. 22:00 & Tue. 02:00 & $l_3$ & $\omega_{30}$ $\omega_{31}$ $\omega_{32}$ \\
\midrule
$g_4$ & Mo. 22:00 & Tue. 02:00 & $l_2$ & $\omega_{33}$ $\omega_{34}$ $\omega_{35}$ $\omega_{36}$ $\omega_{37}$\\
$g_4$ & Tue. 02:00 & Tue. 06:00 & $l_2$ & $\omega_{33}$ $\omega_{34}$ $\omega_{35}$ $\omega_{36}$ $\omega_{37}$ \\
$g_4$ & Tue. 06:00 & Tue. 14:00 & $l_2$ &  $\omega_{2}$ $\omega_{6}$ $\omega_{5}$ $\omega_{15}$ $\omega_{1}$\\

\midrule
$g_5$ & Mo. 06:00 & Mo. 14:00 & $l_2$ & $\omega_{19}$ $\omega_{20}$ $\omega_{21}$ $\omega_{22}$ $\omega_{23}$\\
$g_5$ & Mo. 14:00 & Mo. 22:00 & $l_2$ & $\omega_{21}$ $\omega_5$ $\omega_7$ $\omega_{22}$ $\omega_{1}$\\
\midrule
$g_6$ & Tue. 14:00 & Tue. 22:00 & $l_2$ & $\omega_{17}$ $\omega_{27}$ $\omega_{20}$ $\omega_{19}$ $\omega_{25}$ $\omega_{18}$\\
\midrule
$g_7$ & Tue. 06:00 & Tue. 14:00 & $l_1$ & $\omega_{4}$ $\omega_{12}$ $\omega_{3}$ $\omega_{9}$ $\omega_{8}$  \\
$g_7$ & Tue. 14:00 & Tue. 22:00 & $l_1$ & $\omega_{23}$ $\omega_{26}$ $\omega_{22}$ $\omega_{21}$ $\omega_{16}$\\
\midrule
$g_8$ & Mo. 06:00 & Mo. 14:00 & $l_1$ & $\omega_{10}$ $\omega_{11}$ $\omega_{12}$ $\omega_{13}$ $\omega_{14}$ $\omega_{15}$\\
$g_8$ & Mo. 14:00 & Mo. 22:00 & $l_1$ & $\omega_{24}$ $\omega_{25}$ $\omega_{26}$ $\omega_{27}$ $\omega_{28}$, $\omega_{29}$ \\
$g_8$ & Mo. 22:00 & Tue. 02:00 & $l_1$ & $\omega_{38}$ $\omega_{39}$ $\omega_{40}$ $\omega_{41}$ $\omega_{42}$ $\omega_{43}$ \\
$g_8$ & Tue. 02:00 & Tue. 06:00 & $l_1$ & $\omega_{38}$ $\omega_{39}$ $\omega_{40}$ $\omega_{41}$ $\omega_{42}$ $\omega_{43}$ \\
\bottomrule
\end{tabular*}
\end{table}

\vspace{10cm}
\vspace{10cm}
\vspace{10cm}
\vspace{10cm}
\vspace{10cm}
\vspace{10cm}
\vspace{10cm}
\vspace{10cm}
\vspace{10cm}
\vspace{10cm}
\vspace{10cm}
\vspace{10cm}
\vspace{10cm}
\vspace{10cm}
\vspace{10cm}
\vspace{10cm}
\vspace{10cm}

\subsubsection{Action Space}
\label{sec:action-space}

The action space $\mathcal{A}$ is a finite, discrete set of size
\begin{equation}
|\mathcal{A}| = N_{\text{rows}} \times N_{\text{workers}},
\end{equation}
where $N_{\text{rows}}$ is the number of \textit{line–interval} combinations in the current state, and $N_{\text{workers}}$ is the number of workers under consideration.  
Conceptually, each action corresponds to a tuple
\begin{equation} 
a = (r_{idx}, w_{idx}), \qquad r_{idx} \in \{0, 1, \dots, N_{\text{rows}}-1\}, \quad w \in \{0, 1, \dots, N_{\text{workers}}-1\}
\end{equation}
where $r_{idx}$ denotes the row index (line–interval combination) and $w_{idx}$ the worker index.  
Selecting action $a$ triggers a state transition by updating the allocation flags as follows:
\begin{equation} 
\sigma_{r_{idx},w_{idx}} \leftarrow 1, \qquad 
\sigma_{\bar{r_{idx}},w_{idx}} \leftarrow 0 \quad \forall \: \bar{r} \neq r \ \text{within the same interval}
\end{equation}
ensuring that the worker with index $w_{idx}$ is assigned to exactly one row within the current interval.  
To comply with common RL frameworks, the 2D tuple space is flattened into a single integer index $A$ using lexicographic ordering:
\begin{equation}
A = r \cdot N_{\text{workers}} + w_{idx}, \qquad A \in \{0, 1, \dots, |\mathcal{A}|-1\}.
\end{equation}
The inverse mapping, reconstructing $(r_{idx}, w_{idx})$ from a flattened action index $A$, is given by
\begin{equation}
r = \left\lfloor \frac{A}{N_{\text{workers}}} \right\rfloor, 
\qquad 
w = A \bmod N_{\text{workers}},
\end{equation}
where $\lfloor \cdot \rfloor$ denotes the integer division operator.  
For this mapping to work correctly, rows and workers are indexed starting from $0$.  
In the remainder of this paper, we use 1-based indexing in tables and figures for readability, which introduces a shift of $+1$ when interpreting example values.
In the instance shown in Figure~\ref{fig:mdp-process-example}, there are $N_{\text{workers}} = 43$ and $N_{\text{rows}} = 18$, yielding
\begin{equation}
|\mathcal{A}| = 18 \times 43 = 774 \quad \text{possible actions}.
\end{equation}
The state transition $s_0 \rightarrow s_1$ corresponds to selecting
\begin{equation}
A = 43 \quad \Rightarrow \quad 
(r_{idx}, w_{idx}) = \left(\left\lfloor \frac{43}{43} \right\rfloor, \ 43 \bmod 43 \right) = (1, 0),
\end{equation}
which assigns worker $\omega_1$ (index $w=0$) to row $r=1$.  
Consequently, $\sigma_{1,0}$ is set to $1$ and $\sigma_{0,0}$, $\sigma_{2,0}$ are set to $0$.  
Analogously, the transition $s_{15} \rightarrow s_{16}$ corresponds to
\begin{equation}
A = 171 \quad \Rightarrow \quad 
(r_{idx}, w_{idx}) = \left(\left\lfloor \frac{171}{43} \right\rfloor, \ 171 \bmod 43 \right) = (3, 42),
\end{equation}
assigning worker $\omega_{43}$ (index $w_{idx}=42$) to row $r_{idx}=3$ and updating the corresponding $\sigma$ values.  
Both of these transitions can be observed in Table~\ref{tab:state-space-table} for $k = 43$ workers.

\subsubsection{Reward Function}
\label{sec:rew-function}
In a MDP, the reward function must be shaped such that maximizing the expected cumulative reward corresponds to finding increasingly better solutions to the underlying optimization problem.  
In our case, the reward function combines multiple human-centric factors into a single scalar signal. 
We define the reward function as a weighted sum of three \emph{per-step} human-factor scores and a terminal fairness score:
\begin{align}
R(s, a) &= w_{\pi}\,\pi(s,a) + w_{\rho}\,\rho(s,a) + w_{\xi}\,\xi(s,a) + w_{\mathrm{fair}}^{*}\,f(s'), \label{eq:reward-function}\\[4pt]
f(s') &= 
\begin{cases}
S_{\mathrm{fair}} & \text{if $s'$ is terminal},\\
0 & \text{else}.
\end{cases} \label{eq:fairness-reward}
\end{align}
where $\pi(s,a) \in [0,1]$ denotes the \emph{preference score} of the worker assigned by action $a$ in state $s$, $\rho(s,a) \in [0,1]$ denotes the \emph{resilience score}, i.e., the physical suitability of the worker for the task, $\xi(s,a) \in [0,1]$ denotes the \emph{experience score}, capturing how well-trained the worker is for the assignment, $S_{\mathrm{JS}} \in [0,1]$ is a fairness score computed over the \emph{complete schedule}, as defined below.
The weight parameters $w_{\pi}, w_{\rho}, w_{\xi} \geq 0$ specify the relative importance of the individual human factors, while $w_{\mathrm{fair}} \geq 0$ controls the contribution of the fairness term.  
The terms $\pi, \rho, \xi$ can be evaluated at every environment step, yielding a \emph{dense reward signal} that is well-suited for RL.  
In contrast, fairness can only be meaningfully evaluated once a complete schedule has been produced, because even the final allocation can influence whether the schedule is fair.  
Consequently, $f(s')$ is nonzero only at terminal states, making the fairness signal \emph{sparse}.  
To balance its contribution against the per-step human factors, $w_{\mathrm{fair}}$ must be scaled by the number of allocations $N_{\mathrm{slots}}$ in the episode:
\begin{equation}
    w_{\mathrm{fair}}^{*} = N_{\mathrm{slots}} \cdot w_{\mathrm{fair}}.
\end{equation}
This way the reward $w_{\pi}, w_{\rho}, w_{\xi}, w_{\mathrm{fair}}$ have the same influence on the optimization process (assuming a discount facotor of $\Gamma = 1$). 
For the example shown in Figure~\ref{fig:mdp-process-example}, $N_{\mathrm{slots}} = 83$, meaning $w_{\mathrm{fair}}$ is multiplied by $83$ to make its magnitude comparable to the cumulative dense reward.

We define a schedule as \emph{fair} if workers receive assignments with similar preference levels on average.  
Let $P_i = \{\pi_{i,1}, \pi_{i,2}, \dots, \pi_{i,n_i}\}$ be the set of preference values $\pi$ of worker $\omega_i$ for all tasks they were assigned (rows with $\sigma=1$ in Table~\ref{tab:state-space-table}).  
The mean preference per worker is computed as
\begin{equation}
    \bar{\pi}_i = \frac{1}{n_i} \displaystyle\sum_{k=1}^{n_i} \pi_{i,k}
\end{equation}
where $n_i$ is the number of assignments of worker $\omega_i$.
Let $\mathcal{W}$ denote the set of all workers with at least one assignment ($n_i > 0$).  
The variance of mean preferences across all workers is then
\begin{equation}
    \sigma_{\pi}^2 = \frac{1}{|\mathcal{W}|} \sum_{i \in \mathcal{W}} \left(\bar{\pi}_i - \bar{\pi}\right)^2,
    \qquad
    \bar{\pi} = \frac{1}{|\mathcal{W}|} \sum_{i \in \mathcal{W}} \bar{\pi}_i,
\end{equation}
where $\bar{\pi}$ is the global mean of all workers' mean preferences.
Finally, we normalize this variance to obtain a fairness score $S_{\mathrm{fair}} \in [0,1]$:
\begin{equation}
    S_{\mathrm{fair}} = 1 - \frac{\sigma_{\pi}^2}{\sigma_{\pi,\mathrm{max}}^2},
    \qquad
    \sigma_{\pi,\mathrm{max}}^2 = \frac{1}{4},
\end{equation}
where $\sigma_{\pi,\mathrm{max}}^2 = 1/4$ is the maximum possible variance of a bounded variable in $[0,1]$.  
Hence
\begin{equation}
    S_{\mathrm{fair}} = 1 - 4\sigma_{\pi}^2
\end{equation}
ensuring that $S_{\mathrm{fair}} = 1$ when all workers have identical mean preferences and $S_{\mathrm{fair}} \to 0$ as disparity approaches its theoretical maximum.

\subsection{Implementation}
\label{sec:implementation}
All core business logic is implemented in Python to leverage libraries for optimisation, simulation and machine learning. The first optimisation layer (Order-Line allocation) is implemented using Google OR-Tools' CP-SAT solver \citep{ortools}. 
The CP model encodes setup times, processing durations (dependent on line throughput), machine capacities, and the objective defined in Section~\ref{sec:order-line-def}. 
The solver produces a schedule that is used as input for the second layer.

The second layer is implemented as a Gymnasium environment \citep{towers2024gymnasium} that exposes the MDP described in Sections~\ref{sec:worker-line-def}--\ref{sec:rew-function}. 
The environment implements state construction (tabular representation), action flattening and masking, transition logic, reward calculation and episode termination. 
Action masking is integrated into the environment to prevent selection of infeasible assignments.
This is implemented as a boolean mask returned by the environment together with the action space, in accordance with common RL frameworks (cf. \cite{huang2020closer}).

We support multiple solution strategies. 
RL agents are trained with Stable-Baselines3 \citep{raffin2021stable}, enabling sample-efficient policy learning with masked actions. 
MCTS is provided via the Gymcts library \citep{nasuta2025gymcts}.
The MCTS implementation uses the environment's valid-action mask and the same reward function. 
We also provide greedy allocation approach. 
A manual allocation mode allows experts to step through the environment for validation and demonstration.

The optimization result is serialized as JSON objects, one per line--interval, as illustrated in Listing~\ref{lst:json-example}. 
The \texttt{Start} and \texttt{Finish} fields are UNIX timestamps that define the temporal bounds of the interval. 
The \texttt{Resource} field identifies the production line, while \texttt{geometry} and \texttt{order} identify the produced item and originating order. 
The \texttt{Task} field combines order and geometry identifiers and is produced by the first-layer scheduler. 
The \texttt{produced\_amount} field specifies the quantity to be produced in the interval and \texttt{produced\_until\_now} documents cumulative production to enable progress checks. 
The boolean \texttt{is\_setup\_timebox} marks intervals used exclusively for machine setup. 
The \texttt{workers} array lists unique worker identifiers allocated to the interval. This structured output can be consumed by downstream systems (UI, Orchestrator) for visualization and execution.
\begin{listing}[htb!]
    \caption{JSON representation of a single line–interval assignment, including time bounds, production line, order and geometry identifiers, production progress, setup indicator, and allocated workers. This compact format enables visualization and execution in downstream systems.}
    \label{lst:json-example}
\begin{minted}[frame=single,fontsize=\small,bgcolor=base]{json}
{
  "Finish": 1694510760,
  "Resource": "line 24",
  "Start": 1694493000,
  "Task": "SEV - 38 × 534259180",
  "geometry": "534259180",
  "is_setup_timebox": 0,
  "order": "SEV - 38",
  "produced_amount": 1480,
  "produced_until_now": 6500,
  "required_workers": 4,
  "total_amount": 6500,
  "warning": null,
  "workers": [ "15015261", "15015264", "15015568", "15040627" ]
}
\end{minted}
\end{listing}

The implementation is publicly available in a dedicated GitHub repository (see Appendix~\ref{sec:code-availability}). 
The repository includes examples for solving each layer independently, a complete end-to-end pipeline connecting both layers, preconfigured RL training scripts, and visualization tools for inspecting schedules and allocation trajectories.

\subsection{System Architecture and Integration}
\label{sec:system-architectrue}
This use case is part of the \textit{FAIRWork} EU project, whose overarching goal is the development of a Democratized AI Decision Support System (DAI-DSS). 
The DAI-DSS is designed as a service-oriented, modular architecture that enables the orchestration of complex, interconnected IT systems for production planning and decision support.

\begin{figure}[htb]
    \centering
    \begin{tikzpicture}[scale=0.65, every node/.style={font=\scriptsize}]

    \draw[fill=base] (-1.5,2.5) rectangle ++(3,1.5) node[midway, align=center] {DAI‐DSS \\ User Interface};

    \draw[fill=base] (-1.5,0) rectangle ++(3,1.5) node[midway, align=center] {DAI‐DSS \\ Orchestrator};

    \draw[fill=base] (-1.5,-2.5) rectangle ++(3,1.5) node[midway, align=center] {DAI‐DSS \\ Knowledge base};

    \draw[fill=base] (3.5,0) rectangle ++(3,1.5) node[midway, align=center] {DAI‐DSS \\ AI Enrichment};

    \draw[fill=base] (-6.5,0) rectangle ++(3,1.5) node[midway, align=center] {DAI‐DSS \\ Configurator};

    \draw[latex-latex] (0,1.5) -- ++(0,1);
    \draw[latex-latex] (0,0) -- ++(0,-1);
    \draw[latex-latex] (1.5, 0.75) -- ++(2,0);
    \draw[latex-latex] (-1.5, 0.75) -- ++(-2,0);
    \draw[latex-latex] (1.5, -1.75) -- ++(3.5,0) -- ++(0,1.75);
    \draw[latex-latex] (-1.5, -1.75) -- ++(-3.5,0) -- ++(0,1.75);
        
    \end{tikzpicture}
    \caption{Simplified Overview of DAI-DSS High-Level Architecture of the FAIRWork projekt.}
    \label{fig:sysarchi}
\end{figure}
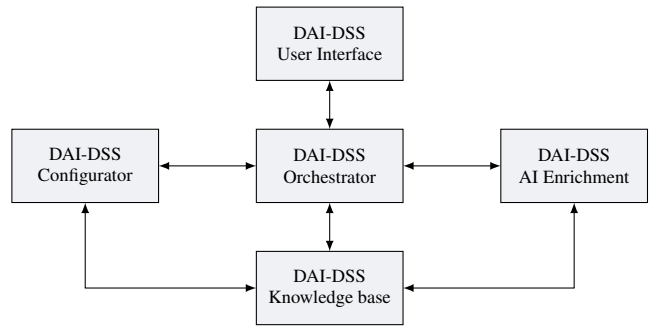

Figure~\ref{fig:sysarchi} illustrates a simplified high-level view of the DAI-DSS architecture as developed within the project. 
The architecture consists of five primary building blocks: 
\textit{DAI-DSS User Interface (UI/UX):} provides the human-facing components of the system, including data upload, visualization, and decision-support dashboards tailored to the use case.  
\textit{DAI-DSS Orchestrator:} coordinates the execution of agents and microservices, managing data flows and triggering optimization workflows.  
\textit{DAI-DSS Configurator:} built on the OLIVE microservice framework, enabling configuration, deployment, and adaptation of solution components.  
\textit{DAI-DSS Knowledge Base:} a central data repository storing static context information such as line capabilities, worker profiles, and historical production data.  
\textit{DAI-DSS AI Enrichment:} a catalogue of AI services that can be invoked for reasoning, optimization, and decision support.  

For the present use case, a web-based user interface was implemented that allows production planners to trigger schedule and worker assignment generation by uploading order data in Excel format. 
This file represents the dynamic input data (cf. Figure~\ref{fig:main-example}). 
Upon receiving input, an \textit{agent} queries the knowledge base for additional context information, such as available production lines and worker profiles, and combines these with the uploaded order data to construct a problem instance. 
This instance is serialized as a JSON document and passed to the service of the  DAI-DSS AI enrichtment component.
Within the Orchestrator, agents  that communicate using Python-based REST endpoints and agent frameworks such as JADE \citep{bellifemine2008jade}. 
The orchestrator coordinates the flow of information between components, triggers AI services, and aggregates results for visualization in the user interface. 
The actual optimization logic, described in Section~\ref{sec:implementation}, is encapsulated as an AI service following the REST paradigm and exposed using the Flask web framework with an OpenAPI/Swagger-compliant interface.
This architecture supports adaptive decision-making by enabling agents to re-trigger optimization workflows when new data becomes available, for example when workers call in sick or production orders change. 
This dynamic re-optimization ensures that production schedules remain feasible and fair even under changing shop-floor conditions.
Detail of the DAI-DSS are discussed in deatil in Projects technical Reports \citep{chevuri2023dai}.

\subsection{Evaluation Setup}
\label{subsec:evaluation_setup}
In order to evaluate the quality and effectiveness of the the service-based approach for scheduling, the schedules proposed by the system are validated regarding technical accuracy and perceived usefulness by domain experts.
We deployed various parametrizations of each layers (different weiths for the objetive or function).
For the first layer we defined the following parametrizations for Equation \ref{eq:objective-def}:
\begin{enumerate}
    \item \textbf{Makespan}: This parametrisazion only optimizes the makespan and disregards tardiness ($w_1c=1, w_\tau=0$)
    \item \textbf{Tardiness}: This parametrization only optimizes the tardiness and disregards tardiness ($w_1=0, w_\tau=1$)
    \item \textbf{Balanced}: This parametrization only optimizes both tardiness an makespan ($w_1=1, w_\tau=1$). One minute of improvement in the makespan reduces the cost or the objective equals one minute of improvenment of the tardiness of task.
\end{enumerate}
For the worker line allocation we can deployed 4 different parameterizations of the reward function (Equation \ref{eq:reward-function}):
\begin{enumerate}
    \item \textbf{Preference}: Maximizes worker satisfaction by taking their preferences for specific tasks in to account. ($w_\pi=1, w_\rho=0, w_\xi=0, w_{fair}=1$).
    \item \textbf{Resilience}: Favors resilience, a for physical and cognitive strain. ($w_\pi=0, w_\rho=1, w_\xi=0, w_{fair}=1$).
    \item \textbf{Experience}: This parametrization focuses on the worker’s proficiency and training level on the specific line and geometry ($w_\pi=0, w_\rho=0, w_\xi=1, w_{fair}=1$).
    \item \textbf{Balanced}: All three human factors experience, preference and resilience are weighted equally.  ($w_\pi=1, w_\rho=1, w_\xi=1, w_{fair}=1$).
\end{enumerate}
Each parametrized reward function was paired with the following solution approaches: 
\begin{enumerate}
    \item \textbf{Greedy}: This approach performs always the action that yields the highest reward.
    \item \textbf{RL}: This trained a poliy $\Pi$ on random problem instances for $500k$ timesteps using Stablebaselines3 the MaskablePPO implementation with the default parameters except for $\Gamma$ wich has been set to 1. This policy is utilized to perform the allocations.
    \item \textbf{MCTS}: This approach use the \textit{Gymcts} library to calculate a solution iterativly. It was set to perform 3 random rollouts per step. That means it generates 3 complete schedules before performing a state transition (for example fro transitioning from $s_0$ to $s_1$ in Figure \ref{fig:mdp-process-example}).
\end{enumerate}
Every possible combination was accessible via a dedicated REST-API endpoint.
In the User Interface the decision maker picks for each of the one possiblity form a dropdown menu and uploads the current order in Excel format. 
The request is passed to the orchestrator where an Agent fetches addtional data from the knowledge base, parses the orders pick the corrresponding endpoint and passes the results back to the UI after the AI server generated a proposed solution.
In total $N_{Total} = 36$ endpoints were deployed.
\begin{align}
    N_{Worker-Line} = N_{Reard\:Function} \times N_{Approach} = 12 \\
    N_{Total} = N_{Order-Line} \times N_{Worker-Line} = 36 
\end{align}
For each generated schedule we calculate a the averade Experiance, preference and resilience scores as KPIs for the schedule.

Through realistically simulated planning scenarios, 10 domain experts evaluate the quality of the proposed solutions.
The validation method includes test sessions, input parameter variation and evaluation questionnaires to collect quantitative scores and qualitative feedback.
Before using the AI system participants received a hands-on training sessions, where the users could familiarize themselves with the interface and the underlying logic of the decision-making model, as well as the evaluation criteria of the system, the configuration parameters and interpretation of the results. 
In total ten test sessions were executed on 16 different weeks.
A test session consists of the following steps:
\begin{enumerate}
    \item The candidate selects a reference week, starting from Monday.
    \item The \textit{days to plan} parameter is set to five days.
    \item The system generates afterwards the schedule with the allocated worker.
    \item The candidate documents the method used for worker allocation.
    \item The candidate assigns a score from 1 to 10 for each generated schedule and worker allocation from the AI system.
    \item Optionally comments could be added for providing qualitative feedback.
    \item The candidate alternates independently configuration parameters on the same reference week, which generates a new schedule using the new parameter configuration.  
    \item All scores are collected and averaged for each week with all comments listed.
\end{enumerate}
The validation questionnaire was structured into two sections: an operational part and a user
perception part.
The first section, called \textit{quantitative test}, focused on operational tasks to measure the effectiveness of the
interface.
The second section collected subjective feedback regarding their experience using the interface.
Although this phase is qualitative in terms of data collection (based on personal perceptions), the evaluation was
carried out using the System Usability Scale (SUS) method, which provides a quantitative score of perceived
usability.
The System Usability Scale (SUS) collects subjective feedback from users, but translates it into a standardized
numerical score, enabling a quantitative evaluation of perceived usability.
Therefore, while the data collection is qualitative in nature (based on personal opinions), the SUS method allows
for a quantification of qualitative data, which justifies the use of the term in the context of this document.
The overall objective is to assess the effectiveness, usability, and intuitiveness of the interface developed for the
Order-Line and Worker-Line allocation capabilities.

\section{Results and Discussion}
\label{sec:results_and_discussion}

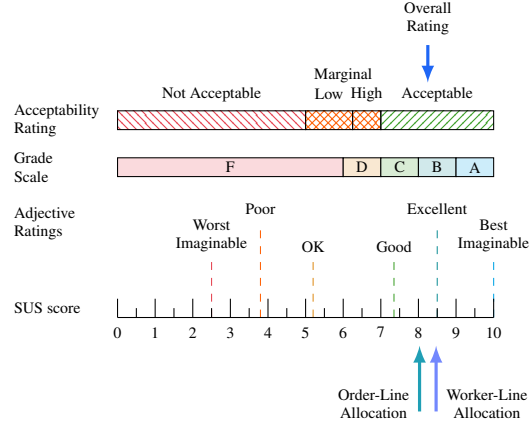
\begin{figure}[htb]
    \centering
\begin{tikzpicture}

\begin{scope}[scale=0.5,]

    \node[align=left, below, anchor=west, font=\tiny] at (-3,2.5) { Adjective \\ Ratings};

    \draw[dashed, draw=maroon] (2.5,0) -- ++(0,1.5) node[above, align=center, font=\tiny]{Worst \\ Imaginable}; 
    
    \draw[dashed, draw=peach] (3.8,0) -- ++(0,2.5) node[above, align=center, font=\tiny]{Poor};
    
    \draw[dashed, draw=yellow] (5.2,0) -- ++(0,1.5) node[above, align=center, font=\tiny]{OK}; 
    
    \draw[dashed, draw=green] (7.35,0) -- ++(0,1.5) node[above, align=center, font=\tiny]{Good}; 
    
    \draw[dashed, draw=teal] (8.5,0) -- ++(0,2.5) node[above, align=center, font=\tiny]{Excellent}; 
    
    \draw[dashed, draw=sky] (10,0) -- ++(0,1.5) node[above, align=center, font=\tiny]{Best \\ Imaginable};

    \node[align=left, below, anchor=west, font=\tiny] at (-3,4.0) {Grade \\ Scale};

    \filldraw[fill=maroon!20] (0,3.75) rectangle ++(6,0.5) node[midway, font=\tiny] {F};
    \filldraw[fill=yellow!20] (6,3.75) rectangle ++(1,0.5) node[midway, font=\tiny] {D};
    \filldraw[fill=green!20] (7,3.75) rectangle ++(1,0.5) node[midway, font=\tiny] {C};
    \filldraw[fill=teal!20] (8,3.75) rectangle ++(1,0.5) node[midway, font=\tiny] {B};
    \filldraw[fill=sky!20] (9,3.75) rectangle ++(1,0.5) node[midway, font=\tiny] {A};

    \node[align=left, below, anchor=west, font=\tiny] at (-3,5.25) {Acceptability \\ Rating};

    \filldraw[pattern={north west lines}, pattern color=maroon, draw opacity=1.0] (0,5.0) rectangle ++(5,0.5);
    \filldraw[pattern=crosshatch, pattern color=peach, draw opacity=1.0] (5,5.0) rectangle ++(1.25,0.5);
    \filldraw[pattern=crosshatch, pattern color=peach, draw opacity=1.0] (6.25,5.0) rectangle ++(0.75,0.5);
    \filldraw[pattern={north east lines}, pattern color=green, draw opacity=1.0] (7,5.0) rectangle ++(3,0.5);

    \node[align=center, above, anchor=south, font=\tiny] at (2.5,5.5) {Not Acceptable};

    \node[align=center, above, anchor=south, font=\tiny, shift={(0,0.035)}] at (5.625,5.5) {Low};
    \node[align=center, above, anchor=south, font=\tiny] at (6.625,5.5) {High};
    \node[align=center, above, anchor=south, font=\tiny] at (6,6) {Marginal};

    \node[align=center, above, anchor=south, font=\tiny] at (8.5,5.5) { Acceptable};

    \draw[latex-, very thick, draw=sapphire] (8.03,-0.75) -- ++(0,-1.75) node[midway, left, align=center, font=\tiny, anchor=north east]{Order-Line \\ Allocation}; 

    \draw[latex-, very thick, draw=lavender] (8.47,-0.75) -- ++(0,-1.75) node[midway, right, align=center, font=\tiny, anchor=north west]{Worker-Line \\ Allocation}; 

    \draw[latex-, very thick, draw=blue] (8.25,6.25) -- ++(0,1.0) node[above, align=center, font=\tiny]{Overall \\ Rating};

    \draw[] (0,0) --  (10,0);

    \foreach \x/\xtext in {0, 1, 2,...,10}
    \draw[shift={(\x,0)}] (0,0) edge (0,0.5) node[align=center, below] {\tiny \x};
    
    \foreach \x/\xtext in {0.5, 1.5,...,10}
    \draw[shift={(\x,0)}] (0,0) -- (0,0.25);

    \node[align=left, below, anchor=west, font=\tiny] at (-3,0.25) { SUS score};

\end{scope}
        
\end{tikzpicture}
    \caption{System Usability Scale (SUS) with reference values and intervals. 
    Arrows indicate the scores achieved for Order–Line, Worker–Line, and overall system evaluation.
    }
    \label{fig:sus-scale}
\end{figure}

The proposed two-layer scheduling system was evaluated in a user study comprising 16 evaluation sessions with domain experts. 
In each session, ten experts assessed both the Order–Line and Worker–Line allocations, thereby providing paired ratings for both tasks per session. 
Figure~\ref{fig:sus-scale} illustrates the SUS rating scale used in the study and summarizes the obtained scores.

The Order–Line allocation achieved an average score of $8.03$ with a standard deviation of $0.64$, while the Worker–Line allocation obtained a slightly higher average of $8.47$ with a standard deviation of $0.92$. 
These results indicate overall satisfaction with both components of the system, with mean scores exceeding eight out of ten. 
The low standard deviations (both below $1.0$) demonstrate consistent performance across a variety of problem instances. 
The higher score of the Worker–Line allocation suggests that experts particularly valued the system’s human-centered worker assignment capabilities. Overall, the results correspond to a SUS evaluation between \textit{good} and \textit{excellent}, leaning toward the latter, but still leaving room for future refinements.

Qualitative feedback from domain experts provided further insights. Regarding the Order–Line allocation, several experts noted that the schedules tended to be very dense, minimizing idle times. 
Although this improves throughput, early completion of production tasks can be disadvantageous, as it may lead to additional storage costs before shipment. 
This behavior is a direct consequence of the objective function defined in Equation~\ref{eq:objective-def}, which penalizes tardiness but does not penalize early completion. As a result, schedules that complete jobs significantly earlier than their deadlines are considered equally optimal. 
A potential improvement would be to replace the tardiness metric with a due-date adherence metric, which penalizes both late and excessively early task completions, thereby aligning production schedules more closely with just-in-time delivery principles.

Another recurring observation concerned seemingly arbitrary idle gaps before certain order-line allocations. 
These gaps are also a consequence of the objective function and the behavior of the CP solver: tasks that are not on the critical path can be shifted in time without affecting the objective value. 
For example, in Figure~\ref{fig:cp-example}, the production of \textit{Geometry~3} could be delayed by up to 1440 minutes without changing the overall objective value. 
The solver likely returns the first feasible solution it encounters, which may contain such idle periods. 
Introducing a due-date adherence objective would mitigate this effect by encouraging task placement closer to ideal start and finish times.

Overall, the generated schedules were perceived as satisfactory. They reliably assigned sufficient workers per shift, respected shift boundaries, and incorporated tool changes without introducing unnecessary idle time. Nevertheless, experts highlighted that within-shift allocations sometimes resulted in unnecessary worker reassignments or splitting operations into multiple identical steps. 
Furthermore, the strict minimization of idle time led to overall early completion of parts, which, as noted earlier, is undesirable because of increased storage costs. 

The quantitative results for each algorithm–strategy combination are summarized in Table~\ref{tab:resource_allocation_results}. The data show that the choice of allocation strategy substantially influences the resulting resource allocation distribution. These effects are further illustrated in the radar plots in Figure~\ref{fig:worker-line-radar-plots}. Each row corresponds to a parameterized reward function, and in every case a clear shift toward the optimized parameter is visible.

\begin{table}[htb]
    \centering
    \caption{Resource allocation results for each algorithm combined with its respective allocation strategy. For each combination, the resulting weight distribution for worker experience, preference, and resilience is reported.}
    \label{tab:resource_allocation_results}

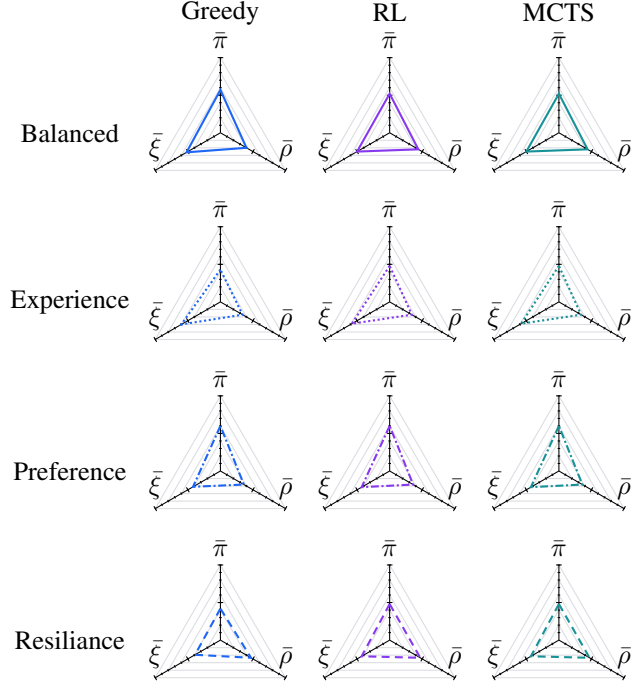
    \caption{Radar plots visualizing Worker–Line allocation KPIs. Each axis is normalized to the range $[0, 1]$. Axis labels are omitted for readability; exact values are reported in Table~\ref{tab:resource_allocation_results}.}
    \label{fig:worker-line-radar-plots}
\end{figure}

The balanced allocation strategy resulted in nearly equal weights for experience, preference, and resilience. 
Strategies focusing on experience or preference caused a marked increase in their respective weights across all three algorithms, accompanied by a decrease in the other weights. 
The resilience-focused strategy increased resilience weights significantly for the greedy and MCTS approaches but less so for RL, where preference weights remained higher than resilience weights. 
This indicates that RL exhibits reduced sensitivity to resilience-focused parameterization, possibly due to the generalization behavior of the trained policy. 
Overall, these findings confirm that the proposed framework can generate resource allocations that explicitly account for worker experience, preference, and resilience, with distinct trade-offs across algorithms.

Additional expert feedback on Worker–Line allocations revealed further areas for refinement. 
The calculation of the fields \texttt{produced\_amount}, \texttt{total\_amount}, \texttt{produced\_until\_now}, and \texttt{is\_setup\_timebox} (cf. Listing~\ref{lst:json-example}, Section~\ref{sec:implementation}) was implemented as a post-processing step. 
As a result, each batch of geometries is split into production and setup intervals. 
During setup intervals, no production occurs, and tooling changes are carried out by specialized staff rather than machine operators. 
Consequently, the MDP process assigned workers to lines during setup intervals, which does not reflect reality. 
This discrepancy slightly affected fairness scores and human factor averages, although its overall impact was negligible since the majority of time is spent in production. 
This modeling limitation could be resolved by incorporating setup intervals directly into the MDP definition as tasks with zero required worker slots.

In some edge cases, the post-processing step also introduced unnecessary task fragmentation, for instance when setup intervals overlapped with shift changes, resulting in additional task splits. 
Furthermore, the current MDP definition allows workers to switch lines when two tasks end simultaneously. 
Experts indicated that in practice, workers should preferably remain on the same line throughout a shift, as reassignments introduce overhead not accounted for in the model. 
This preference could be incorporated by adjusting the allocation logic: instead of reallocating workers only when both line and geometry match, future allocations could prioritize continuity based solely on the line match, thus keeping workers assigned to the same line across consecutive time intervals.

Collectively, these results demonstrate that the two-layer scheduling approach performs well in terms of throughput, fairness, and human-factor considerations, while also revealing specific areas for improvement, most notably the integration of due-date adherence, enhanced continuity of worker assignments, and explicit modeling of setup intervals.

\section{Conclusion}
\label{sec:conclusion}

This work presented a two-layer, human-centric production planning framework that integrates CP for Order–Line allocation with a MDP formulation for Worker–Line allocation. 
The proposed system explicitly incorporates human factors such as worker preference, experience, and resilience into the scheduling process, thereby moving beyond purely efficiency-driven planning towards a more equitable and sustainable production paradigm.

The experimental evaluation, conducted with domain experts from the automotive sector, demonstrated that the approach produces schedules that are both operationally feasible and perceived as fair. 
The Order–Line component reliably generates dense, high-utilization production plans, while the Worker–Line allocation improves perceived fairness and better balances worker assignments compared to baseline methods. 
Expert ratings consistently exceeded eight on a ten-point scale, confirming high overall satisfaction. The relatively low variance across test cases indicates robustness of the approach under diverse production scenarios.

The qualitative feedback highlighted several important insights. First, purely tardiness-based objectives can lead to excessive earliness, suggesting that future research should incorporate due-date adherence or inventory cost considerations. 
Second, explicit modelling of setup intervals within the CP formulation would prevent unintended worker allocations and reduce post-processing complexity. 
Third, continuity constraints or incentives could be introduced to avoid unnecessary mid-shift reassignments, which experts considered disruptive. 
Finally, RL policies exhibited limited sensitivity to reward configurations in some settings, motivating further research into sophisticated training setups.

Future research could extend the framework in several directions. 
Possible avenues include integrating multi-objective optimization to balance earliness, tardiness, and inventory costs, incorporating stochastic processing times and workforce availability to improve robustness, and conducting studies to evaluate the impact of the proposed approach on workforce satisfaction, turnover, and productivity in production environments on a long-term scale.

Overall, the results demonstrate that combining CP-based production scheduling with learning-based worker allocation is a promising direction for advancing human-centric manufacturing systems. 
By providing decision-makers with configurable, transparent, and fair scheduling solutions, the proposed framework contributes to fair workforce management.

\subsubsection*{Author contributions}

\textbf{Alexander Nasuta}: Writing - Original Draft, Writing - Review \& Editing, Conceptualization, Methodology, Software, Visualization \\
\textbf{Alessandro Cisi}: Writing - Original Draft, Writing - Review \& Editing, Conceptualization, Resources, Investigation, \\
\textbf{Sylwia Olbrych}: Conceptualization \\
\textbf{Gustavo Vieira}: Writing - Original Draft, Conceptualization, Software \\
\textbf{Rui Fernandes}: Conceptualization, Software,  Writing - Review\\
\textbf{Marlene Mayr}: Writing - Original Draft, Conceptualization \\
\textbf{Lucas Paletta}: Writing - Original Draft, Conceptualization \\
\textbf{Rishyank Chevuri}: Writing - Original Draft, Data Curation\\
\textbf{Robert Woitsch}: Project administration \\
\textbf{Hans Aoyang Zhou}: Writing - Original Draft, Writing - Review \& Editing\\
\textbf{Anas Abdelrazeq}: Writing - Review \& Editing \\
\textbf{Robert H. Schmitt}: Supervision, Resources \\

\subsubsection*{Acknowledgments}
This work has been supported by the FAIRWork project (\url{www.fairwork-project.eu}) and has been funded within the European Commission’s Horizon Europe Programme under contract number 101069499. 
This paper expresses the opinions of the authors and not necessarily those of the European Commission. 
The European Commission is not liable for any use that may be made of the information contained in this paper.

\subsubsection*{Code Availability}
\label{sec:code-availability}
The source code for the Experiments is available as a \href{https://github.com/Alexander-Nasuta/Optimizing-Fairness-in-Production-Planning}{Github Repository}: \\ \url{https://github.com/Alexander-Nasuta/Optimizing-Fairness-in-Production-Planning}.

\bibliographystyle{apalike}  
\bibliography{references}  

\begin{thebibliography}{}

\bibitem[Alves et~al., 2024]{alves2024sociodemographic}
Alves, J., Lima, T.~M., and Gaspar, P.~D. (2024).
\newblock The sociodemographic challenge in human-centred production systems--a systematic literature review.
\newblock {\em TheoreTical issues in ergonomics science}, 25(1):44--66.

\bibitem[Bellifemine et~al., 2008]{bellifemine2008jade}
Bellifemine, F., Caire, G., Poggi, A., and Rimassa, G. (2008).
\newblock Jade: A software framework for developing multi-agent applications. lessons learned.
\newblock {\em Information and Software technology}, 50(1-2):10--21.

\bibitem[Burgert et~al., 2024]{burgert2024workforce}
Burgert, F.~L., Windhausen, M., Kehder, M., Steireif, N., M{\"u}tze-Niew{\"o}hner, S., and Nitsch, V. (2024).
\newblock Workforce scheduling approaches for supporting human-centered algorithmic management in manufacturing: A systematic literature review and a conceptual optimization model.
\newblock {\em Procedia Computer Science}, 232:1573--1583.

\bibitem[Chevuri, 2023]{chevuri2023dai}
Chevuri, R. (2023).
\newblock Dai-dss architecture and initial documentation and test report.

\bibitem[Finco et~al., 2020]{finco2020considering}
Finco, S., Zennaro, I., Battini, D., and Persona, A. (2020).
\newblock Considering workers’ features in manufacturing systems: a new job-rotation scheduling model.
\newblock {\em IFAC-PapersOnLine}, 53(2):10621--10626.

\bibitem[Gannouni, 2025]{aymenDiss}
Gannouni, A. (2025).
\newblock {\em Reinforcement learning-based optimization of the job shop problem with transportation resources}.
\newblock PhD thesis.

\bibitem[Gerlach et~al., 2024]{gerlach2024exploring}
Gerlach, M., Fabienne, J.~R., Jannic, S.~B., Murat, S., and Golz, C. (2024).
\newblock Exploring nurse perspectives on ai-based shift scheduling for fairness, transparency, and work-life balance.

\bibitem[Huang and Onta{\~n}{\'o}n, 2020]{huang2020closer}
Huang, S. and Onta{\~n}{\'o}n, S. (2020).
\newblock A closer look at invalid action masking in policy gradient algorithms.
\newblock {\em arXiv preprint arXiv:2006.14171}.

\bibitem[Jaehn and Pesch, 2019]{Jaehn2019}
Jaehn, F. and Pesch, E. (2019).
\newblock {\em Ablaufplanung: Einf\"{u}hrung in Scheduling}.
\newblock Springer Berlin Heidelberg.

\bibitem[Jeon et~al., 2016]{jeon2016preferred}
Jeon, I.~S., Jeong, B.~Y., and Jeong, J.~H. (2016).
\newblock Preferred 11 different job rotation types in automotive company and their effects on productivity, quality and musculoskeletal disorders: comparison between subjective and actual scores by workers’ age.
\newblock {\em Ergonomics}, 59(10):1318--1326.

\bibitem[Katiraee et~al., 2021a]{katiraee2021consideration}
Katiraee, N., Calzavara, M., Finco, S., and Battini, D. (2021a).
\newblock Consideration of workforce differences in assembly line balancing and worker assignment problem.
\newblock {\em IFAC-PapersOnLine}, 54(1):13--18.

\bibitem[Katiraee et~al., 2021b]{katiraee2021considerationReview}
Katiraee, N., Calzavara, M., Finco, S., Battini, D., and Batta{\"\i}a, O. (2021b).
\newblock Consideration of workers’ differences in production systems modelling and design: State of the art and directions for future research.
\newblock {\em International Journal of Production Research}, 59(11):3237--3268.

\bibitem[Kellmann and Kallus, 2024]{kellmann2024recovery}
Kellmann, M. and Kallus, K.~W. (2024).
\newblock {\em The recovery-stress questionnaires: A user manual}.
\newblock Taylor \& Francis.

\bibitem[Kemmerling, 2024]{marcoDiss}
Kemmerling, M. (2024).
\newblock {\em Job shop scheduling with neural Monte Carlo Tree Search}.
\newblock PhD thesis.

\bibitem[Kemmerling et~al., 2024]{kemmerling2024beyond}
Kemmerling, M., L{\"u}tticke, D., and Schmitt, R.~H. (2024).
\newblock Beyond games: a systematic review of neural monte carlo tree search applications.
\newblock {\em Applied Intelligence}, 54(1):1020--1046.

\bibitem[Lu et~al., 2022]{lu2022outlook}
Lu, Y., Zheng, H., Chand, S., Xia, W., Liu, Z., Xu, X., Wang, L., Qin, Z., and Bao, J. (2022).
\newblock Outlook on human-centric manufacturing towards industry 5.0.
\newblock {\em Journal of manufacturing systems}, 62:612--627.

\bibitem[Nasuta, 2025]{nasuta2025gymcts}
Nasuta, A. (2025).
\newblock {GYMCTS: A Monte Carlo Tree Search Library for Gymnasium-style Environments (1.2.1)}.
\newblock Version 1.2.1.

\bibitem[Paletta et~al., 2021]{paletta2021towards}
Paletta, L., Ganster, H., Schneeberger, M., Pszeida, M., Lodron, G., Pechst{\"a}dt, K., Spitzer, M., and Reischl, C. (2021).
\newblock Towards large-scale evaluation of mental stress and biomechanical strain in manufacturing environments using 3d-referenced gaze and wearable-based analytics.
\newblock {\em Electronic Imaging}, 33:1--7.

\bibitem[Paletta et~al., 2025]{paletta2025recovery}
Paletta, L., Schneeberger, M., Pszeida, M., Zeiner, H., and Mosbacher, J.~A. (2025).
\newblock Recovery-stress states of workers at the manufacturing site using wearables.
\newblock {\em Cognitive Computing and Internet of Things}, 165:91.

\bibitem[Paletta et~al., 2024]{paletta2024resilience}
Paletta, L., Zeiner, H., Schneeberger, M., Pszeida, M., Mosbacher, J.~A., and Tschuden, J. (2024).
\newblock Resilience scores for decision support using wearable biosignal data with requirements on fair and transparent ai.
\newblock In {\em 2024 IEEE 29th International Conference on Emerging Technologies and Factory Automation (ETFA)}, pages 1--4. IEEE.

\bibitem[Paletta et~al., 2023]{paletta2023digital}
Paletta, L., Zeiner, H., Schneeberger, M., and Quadri, Y. (2023).
\newblock Digital shadows and twins for human experts and data-driven services in a framework of democratic ai-based decision support.
\newblock {\em Cognitive Computing and Internet of Things}, 73(73).

\bibitem[Perron and Furnon, 2025]{ortools}
Perron, L. and Furnon, V. (2025).
\newblock Or-tools.

\bibitem[Raffin et~al., 2021]{raffin2021stable}
Raffin, A., Hill, A., Gleave, A., Kanervisto, A., Ernestus, M., and Dormann, N. (2021).
\newblock Stable-baselines3: Reliable reinforcement learning implementations.
\newblock {\em Journal of machine learning research}, 22(268):1--8.

\bibitem[Rauch et~al., 2020]{rauch2020anthropocentric}
Rauch, E., Linder, C., and Dallasega, P. (2020).
\newblock Anthropocentric perspective of production before and within industry 4.0.
\newblock {\em Computers \& Industrial Engineering}, 139:105644.

\bibitem[Towers et~al., 2024]{towers2024gymnasium}
Towers, M., Kwiatkowski, A., Terry, J., Balis, J.~U., De~Cola, G., Deleu, T., Goul{\~a}o, M., Kallinteris, A., Krimmel, M., KG, A., et~al. (2024).
\newblock Gymnasium: A standard interface for reinforcement learning environments.
\newblock {\em arXiv preprint arXiv:2407.17032}.

\bibitem[Waubert~de Puiseau et~al., 2022]{waubert2022reliability}
Waubert~de Puiseau, C., Meyes, R., and Meisen, T. (2022).
\newblock On reliability of reinforcement learning based production scheduling systems: a comparative survey.
\newblock {\em Journal of Intelligent Manufacturing}, 33(4):911--927.

\end{thebibliography}

\end{document}